\documentclass{article}

\usepackage{PRIMEarxiv}
\usepackage{tikz}
\usepackage[normalem]{ulem}
\usepackage{soul}
\usetikzlibrary{decorations.pathreplacing,fadings}
\usepackage{mathptmx}
\usepackage[utf8]{inputenc} 
\usepackage[T1]{fontenc}    
\usepackage{xcolor}
\usepackage{hyperref}       
\usepackage{url}            
\usepackage{colortbl}
\usepackage{booktabs}       
\usepackage{amsfonts}       
\usepackage{multirow}
\usepackage{nicefrac}       
\usepackage{microtype}      
\usepackage{lipsum}
\usepackage{amsmath}
\usepackage{fancyhdr}       
\usepackage{graphicx}       
\usepackage{orcidlink}
\graphicspath{{media/}}    
\title{Machine Learning for Optical Motion Capture-driven Musculoskeletal Modelling from Inertial Motion Capture Data
\thanks{\textit{\underline{Citation}}: 
\textbf{Dasgupta, A., Sharma, R., Mishra, C., and Nagaraja, V.H. Machine Learning for Optical Motion Capture-driven Musculoskeletal Modeling from Inertial Motion Capture Data. 2023. 1--23.}} 
}
\author{
  Abhishek Dasgupta \orcidlink{0000-0003-4420-0656} \\
Doctoral Training Centre, \\1--4 Keble Road, \\University of Oxford, \\Oxford OX1 3NP, United Kingdom; \\abhishek.dasgupta@dtc.ox.ac.uk \\
   \And
  Rahul Sharma \orcidlink{0000-0003-0700-3098} \\
Laboratory for Computation and Visualization\\in Mathematics and Mechanics,\\ Institute of Mathematics, \\ École Polytechnique Fédérale de Lausanne,\\ Lausanne 1015, Switzerland;\\ rahul.sharma@epfl.ch\\  %% 
   \And
Challenger Mishra \orcidlink{0000-0001-7920-0897} \\
Department of Computer Science \& Technology, \\University of Cambridge, \\15 J.J.\ Thomson Ave., \\Cambridge CB3 0FD, United Kingdom;\\ cm2099@cam.ac.uk\\
   \And
Vikranth H. Nagaraja \orcidlink{0000-0001-7491-8242} \\
 Department of Engineering Science, \\Institute of Biomedical Engineering, \\University of Oxford, \\Old Road Campus Research Building, \\ Oxford OX3 7DQ, United Kingdom;\\ vikranth.harthikotenagaraja@eng.ox.ac.uk \\ 
}

\begin{document}
\maketitle
\vspace{-.3cm}
\begin{abstract}
    Marker-based Optical Motion Capture (OMC) systems and associated musculoskeletal (MSK) modelling predictions offer non-invasively obtainable insights into \emph{in vivo} joint and muscle loading, aiding clinical decision-making. However, an OMC system is lab-based, expensive, and requires a line of sight. 
    Inertial Motion Capture (IMC) systems are widely-used alternatives, which are portable, user-friendly, and relatively low-cost, although with lesser accuracy. 
    Irrespective of the choice of motion capture technique, one needs to use an MSK model to obtain the kinematic and kinetic outputs, which is a computationally expensive tool increasingly well approximated by machine learning (ML) methods.
    Here, we present an ML approach to map experimentally recorded
    IMC data to the human upper-extremity MSK model outputs computed from (`gold standard') OMC input data. 
    Essentially, we aim to predict higher-quality MSK outputs from the much easier-to-obtain IMC data. 
    We use OMC and IMC data simultaneously collected for the same subjects to train different ML architectures 
    that predict OMC-driven MSK outputs from IMC measurements. 
    In particular, we employed various neural network (NN) architectures, such as Feed-Forward Neural Networks (FFNNs) and Recurrent Neural Networks (RNNs) (vanilla, Long Short-Term Memory, and Gated Recurrent Unit) and searched for the best-fit model through an exhaustive search in the hyperparameters space in both \textit{subject-exposed} (SE) and \textit{subject-naive} (SN) settings.
    We observed a comparable performance for both FFNN and RNN models, which have a high degree of agreement (r$_\text{avg, SE, FFNN} = 0.90_{\pm0.19}$, r$_\text{avg, SE, RNN} = 0.89_{\pm0.17}$, r$_\text{avg, SN, FFNN} = 0.84_{\pm0.23}$, and r$_\text{avg, SN, RNN} = 0.78_{\pm0.23}$) with the desired OMC-driven MSK estimates for held-out test data.
    The findings demonstrate that mapping IMC inputs to OMC-driven MSK outputs using ML models could be instrumental in transitioning MSK modelling from `lab to field'.
    \end{abstract}

\keywords{Feed-Forward Neural Network \and  Gated Recurrent Unit \and  Inertial Motion Capture \and  Linear Model \and  Long Short-Term Memory \and  Machine learning \and  Musculoskeletal modelling \and  Optical Motion Capture \and  Recurrent Neural Network \and  Upper Extremity}

\section{Introduction}
    Three-dimensional (3D) motion analysis helps numerically describe joint and segmental kinematics. Besides, it serves to establish non-disabled and identify pathological movement patterns \cite{rau2000movement,Anglin2000}. In recent decades, \emph{in silico} musculoskeletal (MSK) models have offered the ability to non-invasively estimate \textit{in vivo} information to facilitate objective, clinical decision-making to some extent \cite{Erdemir2007,killen2020silico}. In particular, the prediction of intersegmental/joint and muscle loading has proven immensely valuable in improving our understanding of the MSK system \cite{Erdemir2007}.
    However, traditional MSK models can be computationally expensive and typically require extensive input motion capture (mocap) data \cite{Erdemir2007,smith2021review}. In addition, current MSK models can be laborious and too time-consuming to create, scale, and set up for individual subjects. 
    Furthermore, licenses for commercial MSK modelling software can be prohibitively expensive, and some of the open-source packages might be limiting concerning applications and/or features.
Computational complexity and implementation challenges prohibit the routine use of this technique in clinical settings and restrict its use to research environments \cite{Erdemir2007,smith2021review,fregly2021conceptual}. 

A plethora of research and commercial mocap systems are available to help capture 3D human movement data accurately and repeatably \cite{Zhou2008} that can subsequently help describe the MSK function. 
Nonetheless, each mocap system is accompanied by its pros and cons \cite{Zhou2008,Anglin2000}.   
Anglin and Wyss \cite{Anglin2000} noted that stereophotogrammetric (or optoelectronic) systems with passive retro-reflective markers seem to be particularly suited for upper-extremity motion analysis \cite{Cappozzo2005} since they are non-invasive, have high accuracy, and generally do not influence task execution. However, the primary drawbacks of a marker-based Optical Motion Capture (OMC) system include marker occlusion (i.e., line-of-sight requirement), high cost, being lab-based, etc. Conversely, an Inertial Motion Capture (IMC) system is relatively portable and cost-effective, user-friendly, and provides full-body measurement capabilities in the subject’s ecological setting or `field condition'. However, IMC systems might not be as accurate as OMC systems, which are widely regarded as the `gold standard' for non-invasive mocap. Regardless, this `field ready' option of IMC systems can be immensely valuable in low-resource settings and outdoor/real-world applications (e.g., sports, ergonomics). 
Wearable inertial sensors have evolved rapidly and are routinely used in different areas of clinical human movement analysis, e.g., gait analysis \cite{jung2022applied}, stabilometry, instrumented clinical tests, upper-extremity mobility assessment, daily-life activity monitoring, and tremor assessment \cite{iosa2016wearable}. Such sensors have undergone a rapid transition from use in laboratory-based clinical practice to unsupervised, applied settings \cite{picerno2021wearable}.

    There is a growing need to democratise and take objective MSK model-based analysis from `lab to field' \cite{verheul2020measuring,saxby2020machine,arac2020machine,frangoudes2022assessing}. Inverse-dynamics-based MSK models have traditionally been driven by OMC data \cite{Damsgaard2006,Delp2007}); however, recent studies have demonstrated the use of IMC input for MSK models \cite{karatsidis2019musculoskeletal, Konrath2019,nagarajamarker,larsen2020estimation,skals2021manual2,di2022inertial}. %Karatsidis2017
   Notwithstanding, research on upper-extremity biomechanics has lagged compared to that involving the lower extremity \cite{rau2000movement,Anglin2000,philp2022international}   due to several barriers such as a lack of consensus on a standardised protocol, functional tasks, model complexity, tracking method, marker protocol, and performance metrics. 
 Thus, it is imperative to undertake better modelling of upper-extremity biomechanics as they are integral to many activities of daily living.

Machine learning (ML) has received considerable attention in human motion analysis, especially in the last decade \cite{halilaj2018machine,saxby2020machine,xiang2022recent,arac2020machine,frangoudes2022assessing}.
 In particular, using supervised ML models to bypass computationally expensive biomechanical models has become commonplace.
Numerous studies have reported ML applications for estimating \emph{in vivo} joint and muscle loading information from either OMC or IMC data \cite{amrein2023machine,lee2022inertial,sharma2022machine,wouda2018estimation,fernandez2020population,Sohane2020,mubarrat2022shoulder,cleather2019neural,giarmatzis2020real,mundt2020estimation,mundt2020prediction}. Among ML approaches, deep learning models, including Convolutional Neural Networks (CNN) and, increasingly, Recurrent Neural Networks (RNN), have become popular in lower-extremity biomechanical analyses \cite{xiang2022recent}.
Such ML methods have multiple advantages over their MSK model-based counterparts -- (i) while the initial training of the ML model
takes time, prediction corresponding to new data is computationally efficient, facilitating real-time applications; and (ii) they do not require the large amounts of data that are required for MSK models and help provide population-level insights \cite{fernandez2020population}.

Our previous work \cite{sharma2022machine} estimated upper-extremity MSK model information solely from IMC input data. 
This study uses ML methods to map IMC input data to OMC-driven MSK outputs.
It serves two purposes: (i) bypasses the dependency on the computationally expensive MSK model during the prediction phase;
\textit{and}, (ii) more importantly, obviates the requirement for an OMC system in the prediction phase but still estimates `OMC-quality' MSK outputs.
It allows the prediction of (non-invasive) gold-standard OMC-quality MSK outputs for mobile/ambulatory applications (e.g., sports, ergonomics, recreational activities) or in low-resource settings using highly portable IMC systems. 
A similar study for the lower extremity has used simulated IMU inputs \cite{mundt2020predictionimcomc}. 
This work differs in using actual IMC inputs, which were synchronously collected with the OMC inputs, giving greater reliability in predicting OMC quality outputs and in its application to the upper extremity. 
Like that study, we have realised our objective using an ML approach.
In particular, we used Feed-Forward Neural Networks (FFNN) and RNN to model the human upper-extremity motion analysis.
RNNs are, in general, better than FFNNs for modelling time-series data, and in this work, we have explored the popular RNN cells (vanilla, Long Short-Term Memory [LSTM], and Gated Recurrent Unit [GRU]) in RNNs and bidirectional RNNs (in which the input time-series is processed in both forward and backward direction) networks.
It is important to note that such a model creation was only possible because of the unique dataset acquired for \cite{nagarajamarker}, where the upper-extremity motion for five non-disabled subjects was synchronously captured using both IMC and OMC systems.
To the best of our knowledge, this is the first study (across upper and lower extremities) that uses the comparatively lower-quality (but highly portable) IMC data to estimate kinematic and kinetic variables on par with the gold standard OMC-driven MSK outputs.
Such an approach will be instrumental in getting the technology from `lab to field' where OMC-driven systems are infeasible.

Finally, along with an entirely novel approach to map IMC inputs to OMC-driven MSK outputs, this study also addresses several other research gaps in this field.
For instance, most research in this field lacks rigorous hyperparameter tuning when training ML architectures (highlighted in \cite{halilaj2018machine} and \cite{sharma2022machine}) which is crucial for generalisable models.
Moreover, we trained the selected ML models for two different settings -- (i) \textit{subject-naive} (all trials from a subject are strictly either in test or training data), and (ii) \textit{subject-exposed} (at least one trial from each subject is used in the training of the model). 
The \textit{subject-naive} setting is the true check for the generalisability of ML models, which is often ignored.
Moreover, from the motion analysis point of view, this study focuses on modelling the human upper extremity, a relatively under-studied but essential domain.

\section{Materials and methods}

\subsection{Data}
\label{sec:Data}    
    The anonymised dataset used here is from an earlier study \cite{nagarajamarker}, approved by local Research Ethics Committees (Ref no. 16/SC/0051 and Ref no. 14/LO/1975). 
    The study involved five male adult non-disabled volunteers (Age: 22.80 $\pm$ 0.84 years; Height: 1.75 $\pm$ 0.07 m; Weight: 66.25 $\pm$ 9.72 kg; Body Mass Index (BMI): 21.79 $\pm$ 3.49 kg/m$^2$) had consented to participate.

\begin{figure*}[htp]
        \centering
        \includegraphics[width=\textwidth]{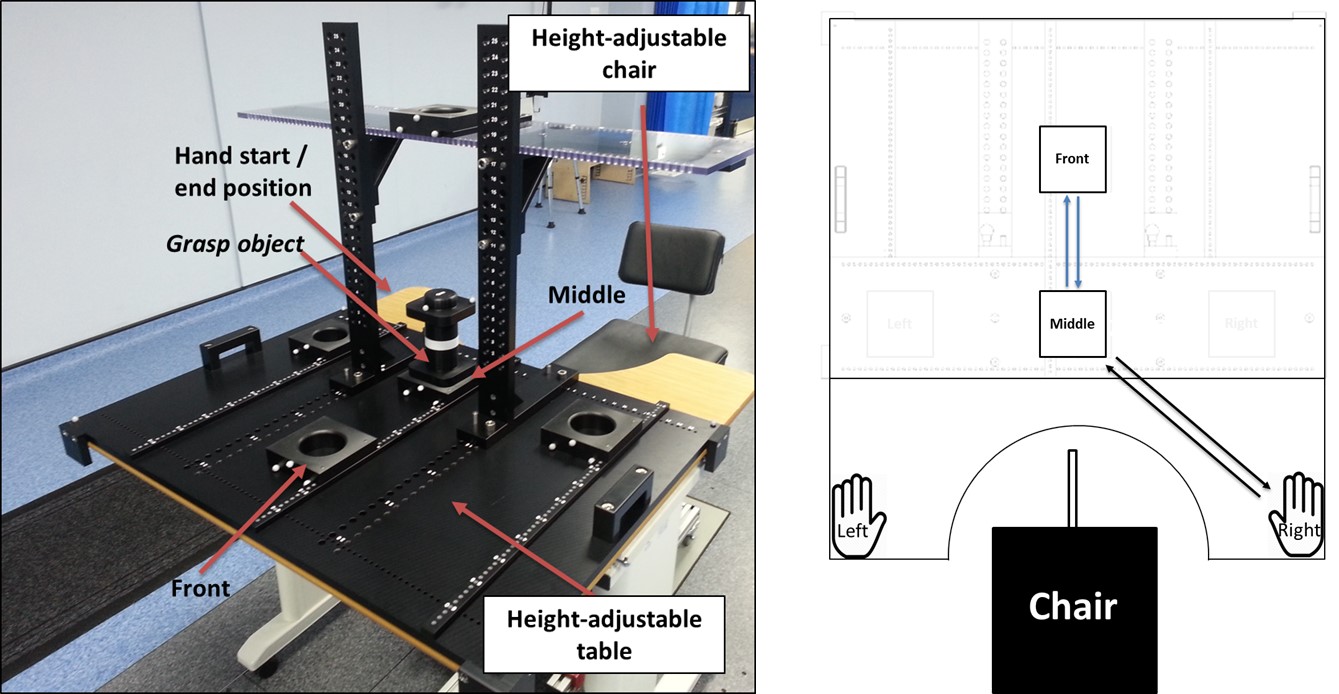}
        \caption{Left side: Custom-built apparatus for \textit{Reach-to-Grasp} task execution in the Forward direction; Right-side: Reach-to-Grasp task setup.
        }
        \label{fig:Figure_1}
    \end{figure*}

    For each subject, three trials of the Reach-to-Grasp task in the \emph{Forward} direction were executed at a self-selected pace using a custom-built apparatus (Supplementary Figure~\ref{fig:Figure_1}) to accommodate the anthropometric requirements of different participants and facilitate the execution of the \textit{Reach-to-Grasp} task in a seated position.
    The task involved the participant reaching to grasp a  
    `dumbbell-shaped' object and moving it between various pre-defined points as instructed. 
    The pre-defined \textit{‘Front’} point was within 90\% of an individual’s arm length (i.e., acromion to middle-fingertip length, with the arm hanging down) to minimise the contribution of trunk movement for task execution \cite{nagaraja2018compensatory,nagaraja2023comparison}.
    
    Synchronised data capture corresponding to task execution involved a 16-camera Vicon\textsuperscript{TM} T40S Series System (Vicon Motion Systems, Oxford, UK) for OMC input data (sampling frequency: 100 Hz) and a Full-body Xsens MVN Awinda Station with 17 wireless IMU sensors (Xsens Technologies B. V., Enschede, The Netherlands) for the IMC input data (sampling frequency: 60 Hz). Hence, we down-sampled the OMC-driven MSK output using cubic interpolation to match IMC input data's sampling frequency ($\sim 60$ Hz).  
    Passive retro-reflective markers (ø 9.5 mm) were placed on the participant’s skin using a double-sided hypoallergenic tape using the Plug-in Gait marker set \cite{PluginGa94:online}. Each IMU contains a 3D accelerometer, gyroscope and magnetometer, along with a barometer and thermometer, and transmits data wirelessly to a master receiver in real-time \cite{paulich2018xsens}. IMU sensors' placement on the participants’ body segments and their setup and calibration was based on the recommended protocol from the Xsens MVN User Manual \cite{MVNUserM34:online}.
    Data capture was synchronised between the two systems wherein the Xsens IMC system triggered the Vicon OMC system following the guidelines of Xsens with a specific trigger at the start/stop recording time \cite{Synchron68:online}. Marker and sensor placement, as well as data capture for all subjects, were performed by the same tester (VHN) to remove inter-tester variability error.
    
    OMC data was processed using Vicon Nexus\textsuperscript{TM} v.2.5 software \cite{PDFdownl80:online} to clean and label the marker trajectories for MSK model inputs. The processed marker trajectories were exported in Coordinate 3D (.C3D) file format \cite{C3DORGTh92:online}. Meanwhile, the affiliated Xsens MVN Analyze 2018 software \cite{MVNAnaly30:online} was used to capture the IMC data and export it as BioVision Hierarchy (.BVH) files for MSK model inputs where a stick-figure model was initially reconstructed. 
    Upper-extremity MSK modelling was carried out using AnyBody\textsuperscript{TM} Modeling System (AnyBody\textsuperscript{TM} Technology A/S, Aalborg, Denmark) separately for each of the mocap inputs as well as subject-specific anthropometric dimensions for scaling following these sources \cite{harthikote2019motion,nagaraja2023comparison,karatsidis2019musculoskeletal}.
    The \emph{`Simple Full-Body model'} and \emph{`Inertial MoCap model'} in the AnyBody Managed Model Repository (AMMR) v.2.1.1 were adapted separately to calculate the respective kinematic variables (i.e., joint angles) and kinetic variables (i.e., joint reaction forces, joint moments, and muscle forces) of interest.

    \begin{figure*}[htb]
        \centering
        \includegraphics[width=\textwidth]{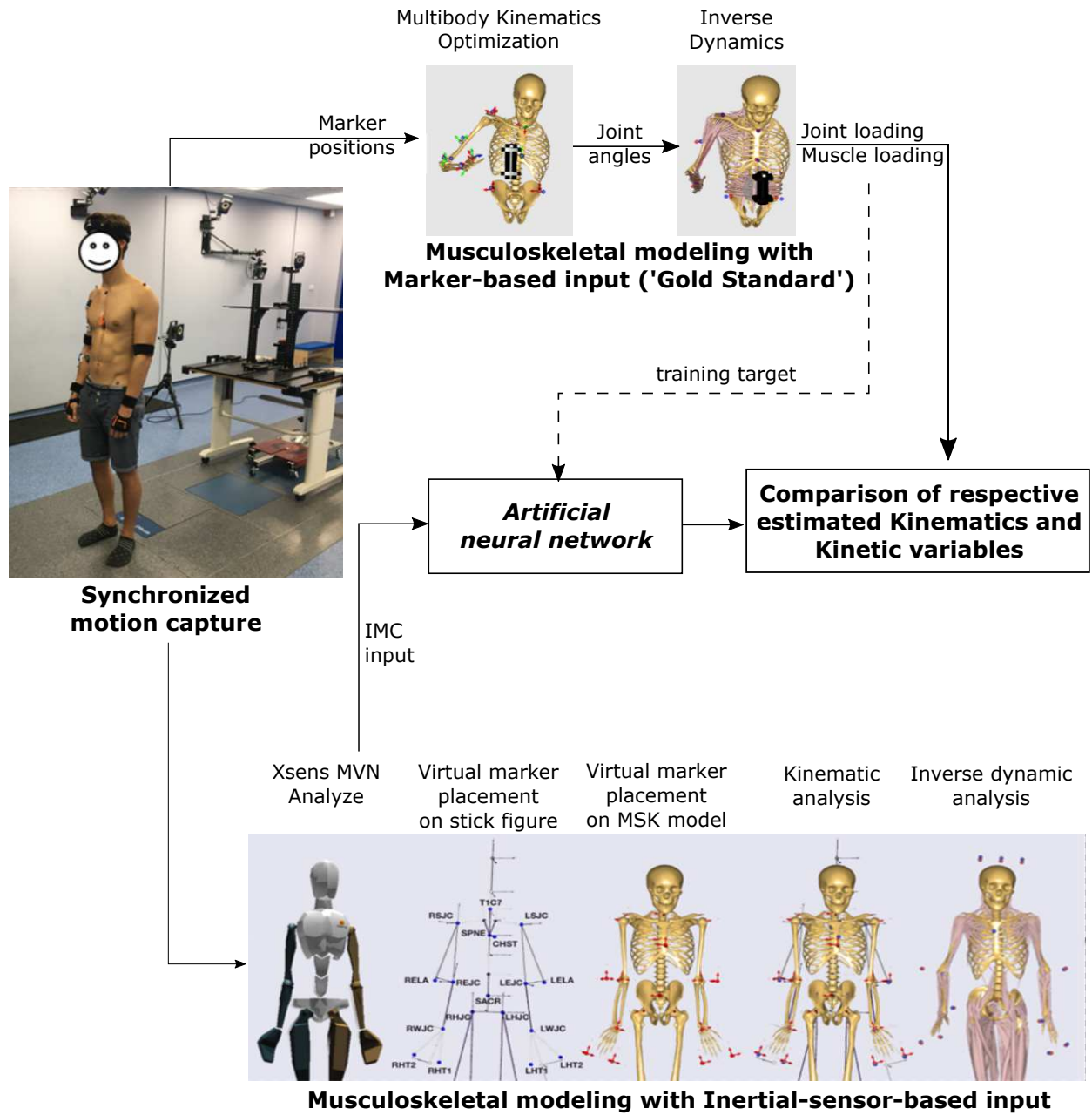}
        \caption{Pipeline for machine learning (ML) for musculoskeletal (MSK) modelling combining the marker-based (or Optical Motion Capture) and inertial-sensor-based (or Inertial Motion Capture) data; Image adapted from \cite{harthikote2019motion,karatsidis2019musculoskeletal,nagarajamarker}
        }
        \label{fig:Pipeline}
    \end{figure*}

    Demographic and anthropometric characteristics (e.g., age, height, weight, gender) were found to influence the amplitude of kinematic and kinetic variables in clinical movement analysis, and if not corrected, individual differences may act as confounding factors \cite{moisio2003normalization,derrick2020isb}. Therefore, the kinetic outcome variables were reported per the recommendations by the \textit{International Society of Biomechanics} on reporting intersegmental forces and moments during human motion analysis \cite{derrick2020isb}. Here, the joint reaction force and muscle force values were normalised to the corresponding subject's body weight and were expressed as a percentage of body weight (\%Body Weight [\%BW]) to facilitate inter-individual comparison \cite{Bergmann2007}. The joint moment values were normalised to the corresponding subject's body weight times height and were expressed as a percentage of body weight times body height (\%Body Weight $\times$ Body Height [\%BW$\times$BH]), since the `body weight times height' normalisation method was found to be more effective in reducing differences primarily due to gender than the `body weight' normalisation method \cite{moisio2003normalization}. The sign convention adopted for the joint angles was -- Flexion (Fl), Forward bending (Fb), Abduction (Ab), Right bending (Rb), Radial deviation (Rd), and Internal rotation (Int) are positive; Extension (Ex), Backward bending (Bb), Adduction (Ad), Left bending (Rb), Ulnar deviation (Ud), and External rotation (Ext) are negative \cite{nagaraja2023comparison}.

   Detailed instructions and a couple of practice sessions were provided for each subject prior to data collection. First, a calibration sequence was performed for the IMC mocap system, which involved the participants standing in a neutral posture (`N-pose') and walking a few steps forward and back to the starting position. Next, participants performed a ‘Static’ trial by standing in a stationary neutral or anatomical position in the centre of the capture volume for the OMC system. This was followed by three trials of the Reach-to-Grasp task with the subject's dominant hand at a self-selected pace. Execution of the task involved moving the hand from the \textit{‘Start/End’} position to grasp the `dumbbell-shaped' object from the \textit{‘Middle’} block and placing it on the \textit{‘Front’} block, then without letting go of the object, the hand would follow the same instructions/path in the return phase, and then the hand would return to the \textit{‘Start/End’} location.
    In contrast with an earlier similar study solely involving MSK modelling with OMC input \cite{harthikote2019motion}, the weight of the grasped object (`dumbbell') was considered to be a point object with negligible mass in the current MSK output involving the IMC or OMC data \cite{nagarajamarker}. Notably, this assumption does not affect the objectives of the current study.
    The experimental data captured from the IMC system and the OMC-data-driven MSK model estimations (i.e., kinematic and kinetic parameters) were used as training data for our ML endeavours.

\subsection{Supervised learning} 
    
   We developed ML models to obtain OMC-driven MSK outputs from IMC inputs for the human upper extremity (Figure \ref{fig:Pipeline}).  
   The IMC experimental data comprise 15 trials (i.e., three trials of Reach-to-Grasp task each obtained from the five subjects) with 483 input features \cite{MVNUserM34:online} while the five OMC-driven MSK model output categories comprise time-series of ten joint angles, twelve joint reaction forces, ten joint moments, four muscle forces, and four muscle activations. We train different ML models for each output category while using IMC experimental data as input. 
    We apply the following feature transformations, similar to those in our previous article \cite{sharma2022machine} before training the model:
    
    \begin{enumerate}
\item The time column, normalised between zero and one, represents the fraction of the total time elapsed for completing the task at a self-selected pace. This helps capture the temporal aspect of MSK analyses.
\item Muscles are typically discretised into numerous muscle bundles in MSK models. MSK model outputs for muscle forces and muscle activations comprise features for the four considered superficial muscle groups, i.e., (a) Pectoralis major (Clavicle part); (b) Biceps Brachii; (c) Deltoid (Medial); and (d) Brachioradialis \cite{harthikote2019motion}. 
The `maximum envelope' of the tendon forces of selected bundles constituting a particular muscle was calculated for further analysis. Muscle activation is a measure of the force in a selected muscle relative to its strength. 
\item We have used many-to-one RNN architecture, which uses multiple previous inputs for an output; therefore, we transformed the time-series data into a sub-time-series of $t$ frames by sliding across the original time-series in a step of one. Thus, the input of RNN is $t \times 483$, where $t$ was taken as 10.
    \end{enumerate}
   
To show the added utility of applying complex ML techniques, we first attempt a linear model to see if it provides satisfactory metrics.
We use Pearson's correlation coefficient (r) and Normalised root-mean-squared error (NRMSE) metrics to compare the effectiveness of our models. 
For \textit{subject-exposed} (SE) models, linear models perform `moderately' to `strongly' in terms of r. For \textit{subject-naive} (SN) models,
linear models perform moderately in terms of r; however, the NRMSE is large for SN models (as shown in Supplementary Table~\ref{tab:result1} as well as Sections \ref{sec:result} and \ref{sec:discuss}). Therefore, it was concluded that complex ML models are necessary.

    \begin{figure*}[htb]
        \centering
    \includegraphics[width=0.62\textwidth]{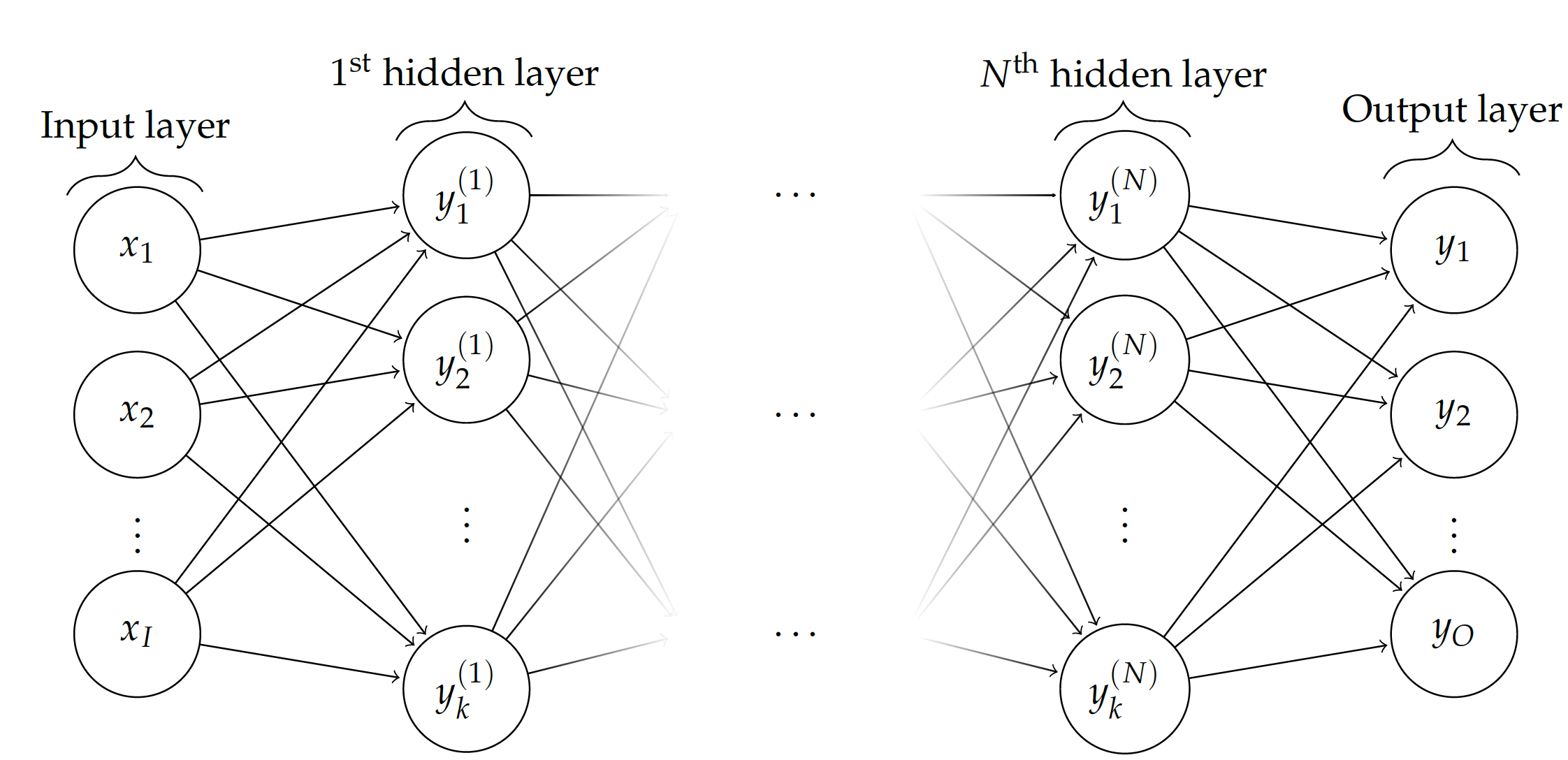}
	\caption{Typical schematic diagram for a Feed-forward Neural Network (FFNN) with $I$ input units, $O$ output units, and $N$ hidden layers each containing $k$ neurons.
	The input and output layers are considered as $0^\text{th}$ and $(N+1)^\text{th}$ layers. 
	}
    \label{fig:NN}
    \end{figure*}
    
    \begin{figure*}[htb]
        \centering
        \includegraphics[width=\textwidth]{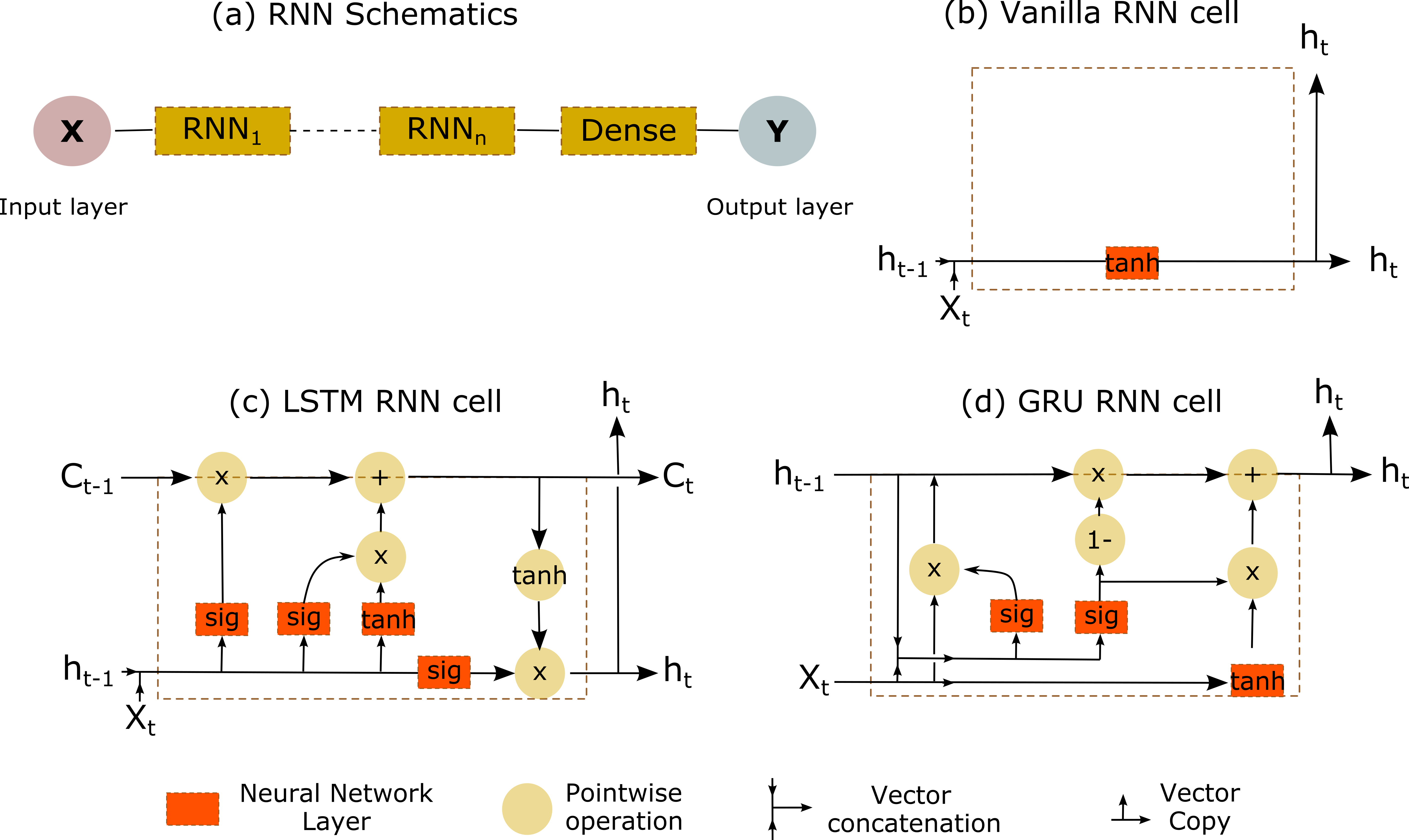}
        \caption{(a) Typical deep Recurrent Neural Network (RNN) architecture containing $n$ RNN cells and a final dense layer \cite{Goodfellow-et-al-2016,hochreiter1997long,cho2014properties}. The schematics of various RNN cells considered include: (b) vanilla RNN cell, (c) Long Short-Term Memory (LSTM) cell, and (d) Gated Recurrent Unit (GRU) cell. Note: $X_t$ is input at any time $t$, $C_t$ is cell state, and $h_t$ is hidden state.
        Further note that the default activation function in RNN cells is tanh while for recurrent steps in LSTM and GRU cells, the default recurrent activation function is sigmoid (sig).
        }
        \label{fig:RNN}
    \end{figure*}

    Two neural network architectures were employed with normalised MSE as the cost function: (i) FFNN, depicted in Figure~\ref{fig:NN} shows a typical schematic diagram, with $I$ input units, $O$ output units, and $N$ hidden layers with $k$ neurons each, with No activation function for the final layer, and
    (ii) RNN, depicted in Figure~\ref{fig:RNN} shows a schematic architecture (containing at least one RNN cell) along with choices of RNN cells, including vanilla, LSTM, and GRU in both unidirectional and bidirectional (which processes the input time-series in both forward and backward directions) sense \cite{Goodfellow-et-al-2016,hochreiter1997long,cho2014properties}.
    Here, it is worth noting that we have treated these choices in RNN cells as hyperparameters. 
    
    Then the validation accuracy was used to tune FFNN and RNN hyperparameters through an exhaustive grid search in the space with 43,740 and 23,328 choices that include hyperparameters such as weight initialization, optimizer, batch-size, epoch, activation function, number of nodes, hidden layers, learning rate, RNN cells, and dropout probability, as listed in Tables \ref{tab:NN_param} and \ref{tab:RNN_param} along with the optimal hyperparameters.
    The current hyperparameter exploration for FFNNs is informed by our previous related work \cite{sharma2022machine}.

\begin{table}[htp]
\vspace{1cm}
\begin{center}
\scalebox{0.8}{
\begin{tabular}{ c c c c c c c c c c }
\hline
 Output & Weight  & Optimizer  & Batch  & Epoch & Activation  & Number   &  Hidden & Learning & Dropout   \\  
   &  initialization & &  size &  &  function & of nodes  &  layers & rate & probability \\
 \hline
 \hline
\vspace{-0.2cm}
&  &  &   &   &   &   &  &  &  \\
 \multicolumn{10}{c}{\textbf{Hyperparameters space explored}}  \\
\vspace{-0.2cm}
&  &  &   &   &   &   &  &  &  \\
    & Xavier normal,  & Adam,   & 64,   & 50,  & ReLU,    & 200 to 1800 &  2, 4,  & 0.001, & 0,   \\
  & Random normal,  & SGD,    & 256,  & 100, & tanh,    & in steps of 200 & 6, 8, 10  & 0.005 & 0.2   \\
       & He normal      & RMSProp & 1028 & 200 & sigmoid & &  & &   \\
\hline
\multicolumn{10}{c}{} \\[-0.7em]
\multicolumn{10}{c}{\textbf{Optimal hyperparameters in \textit{subject-exposed}  settings}}   \\
\multicolumn{10}{c}{} \\[-0.7em]
Joint angles          & Random normal & Adam & 256  & 200 & ReLU    & 200 & 2 & 0.001 & 0.0 \\
Joint reaction forces & Xavier normal     & Adam & 256  & 100  & ReLU & 200 & 8 & 0.005 & 0.0 \\
Joint moments         & Xavier normal & Adam & 64   & 50  & sigmoid & 1400 & 2 & 0.001 & 0.2 \\
Muscle forces        & Random normal & SGD & 64 & 50 & ReLU & 400  & 8 & 0.005 & 0.0 \\
Muscle activations        & Xavier normal & Adam & 256 & 100 & ReLU & 1000  & 6 & 0.005 & 0.0 \\
\hline
\vspace{-0.2cm}
&  &  &   &   &   &   &  &  &  \\
 \multicolumn{10}{c}{\textbf{Optimal hyperparameters in \textit{subject-naive}  settings}}    \\
\vspace{-0.2cm}
&  &  &   &   &   &   &  &  &  \\
Joint angles   & Random normal & Adam & 256   & 100 & ReLU    & 800 & 4 & 0.005 & 0.0 \\
Joint reaction forces & Random normal & Adam & 256  & 50  & ReLU    & 800 & 8 & 0.001 & 0.0 \\
Joint moments          & Xavier normal & Adam & 64  & 50 & sigmoid &  1800 & 2 & 0.001 & 0.2 \\
Muscle forces      & Xavier normal & SGD & 64 & 200  & ReLU & 1200 & 8 & 0.005 & 0.2 \\
Muscle activations        & Random normal & Adam & 256 & 200  & ReLU & 1200 & 6 & 0.001 & 0.2 \\
\hline
\end{tabular}
}
\end{center}
\caption{Hyperparameters choices explored (43,740) for Feed-Forward Neural Network (FFNN) along with optimal hyperparameters for each output category in \textit{subject-exposed}  and \textit{subject-naive}  settings.
}
\label{tab:NN_param}
\end{table}

\subsection{Validation and train-test split}

    Two approaches are commonly used to benchmark ML architectures that approximate MSK models: \textit{subject-exposed} (SE) and \textit{subject-naive} (SN) \cite{rane2019deep}. 
    The trials of all subjects are pooled together in the SE setting, and a train-test split is performed on the trials, ensuring that each subject in the test set has at least one trial in the training set. 
    An ML model trained using this approach usually performs better than the SN setting, where a subject is strictly either in training or test data. While the performance is better, the SE model has the disadvantage of being a subject-specific model and does not generalise well \cite{giarmatzis2020real}. 
    
    In the train-test split for the SE case, two trials were randomly selected as the test set and the remaining trials as the training set.
    For hyperparameter tuning, we performed cross-validation by randomly 
    splitting the remaining 13 trials into validation (two trials) and training sets (11 trials).
    While in the SN case, one of the subjects was randomly selected as the test subject and the remaining
    subjects as the training data. Subsequently, cross-validation was performed by randomly selecting one subject as the validation data and the remainder as the training data. 

\begin{table}[htp]
\begin{center}
\scalebox{0.85}{
\begin{tabular}{ c c c c c c c c c c }
\hline
 Output & RNN cell & Optimizer  & Batch  & Epoch & Activation  & Number   &  RNN & Learning & Dropout   \\  
   &   & &  size &  &  function & of nodes  &  layers & rate & probability \\  
 \hline
 \hline
\vspace{-0.2cm}
&  &  &   &   &   &   &  &  &  \\
 \multicolumn{10}{c}{\textbf{Hyperparameters space explored for RNN}}  \\
\vspace{-0.2cm}
&  &  &   &   &   &   &  &  &  \\
     & Vanilla, LSTM,   & Adam,    & 64,   & 50,  & ReLU,    & 128, &  1, 2,  & 0.001, & 0.1,   \\
     & GRU, B-Vanilla,  & SGD,     & 128,  & 100, & tanh,    & 256, & 3, 4    & 0.005 & 0.2   \\
     & B-LSTM, B-GRU    & RMSProp  & 256   & 200  & sigmoid  & 512  &         & &   \\
\hline
\vspace{-0.2cm}
&  &  &   &   &   &   &  &  &  \\
 \multicolumn{10}{c}{\textbf{Optimal hyperparameters in \textit{subject-exposed}  settings}}   \\
\vspace{-0.2cm}
&  &  &   &   &   &   &  &  &  \\
Joint angles & BLSTM & Adam & 256 & 200 & sigmoid & 256 & 1 & 0.001 & 0.2 \\
Joint reaction forces & LSTM & RMSprop & 128 & 50 & sigmoid & 128 & 2 & 0.001 & 0.1 \\
Joint moments & BLSTM & RMSprop & 64 & 100 & sigmoid & 256 & 1 & 0.001 & 0.1 \\
Muscle forces & LSTM & RMSprop & 64 & 100 & tanh & 256 & 1 & 0.001 & 0.1 \\
Muscle activations & LSTM & RMSprop & 64 & 50 & sigmoid & 128 & 3 & 0.001 & 0.1 \\
\hline
\vspace{-0.2cm}
&  &  &   &   &   &   &  &  &  \\
 \multicolumn{10}{c}{\textbf{Optimal hyperparameters in \textit{subject-naive}  settings}}    \\
\vspace{-0.2cm}
&  &  &   &   &   &   &  &  &  \\
Joint angles & BLSTM & Adam & 64 & 50 & tanh & 256 & 1 & 0.001 & 0.2 \\
Joint reaction forces & LSTM & Adam & 64 & 50 & relu & 128 & 4 & 0.001 & 0.1 \\
Joint moments & GRU & Adam & 128 & 100 & sigmoid & 256 & 2 & 0.005 & 0.2 \\
Muscle forces & GRU & Adam & 256 & 100 & relu & 512 & 3 & 0.001 & 0.2 \\
Muscle activations & LSTM & RMSprop & 128 & 100 & tanh & 512 & 2 & 0.001 & 0.1 \\
\hline
\end{tabular}
}
\end{center}
%\end{adjustwidth}
\caption{Hyperparameters choices explored (23,328) for the Recurrent Neural Network (RNN) along with optimal hyperparameters for each output category in \textit{subject-exposed}  and \textit{subject-naive}  settings. In the RNN cell category, `B' stands for Bidirectional cell (which processes the input time series in both forward and backward directions), LSTM is a Long Short-Term Memory cell, and GRU is a Gated Recurrent Unit cell.
    }
\label{tab:RNN_param}
\end{table}

\subsection{Error metrics}\label{sec:metric}
    NRMSE (i.e., RMSE scaled by the standard deviation of the variable in the dataset) was used as our primary error metric as in \cite{sharma2022machine,giarmatzis2020real}, where the ground truth is the corresponding OMC-driven MSK output.
    Since RMSE (and thus NRMSE) is sensitive to outliers, we also used Pearson's correlation coefficients (r) between the MSK model outputs and the ML model-predicted outputs as a secondary measure. 

\section{Results}\label{sec:result}
\begin{figure*}
        \centering
        \includegraphics[width=\textwidth]{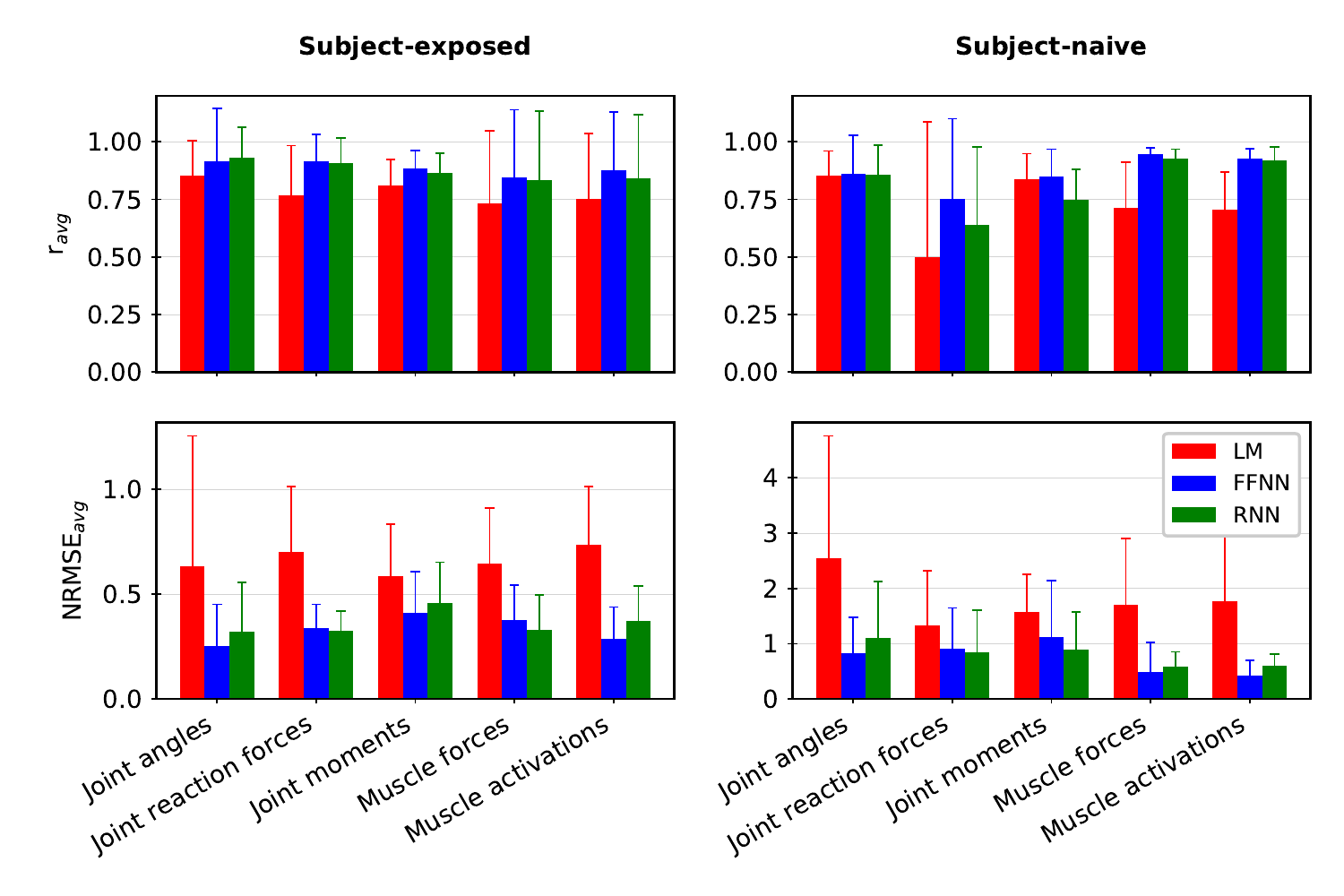}
        \caption{
            Average Pearson's correlation coefficient (r$_{\text{avg}}$) and average NRMSE values (NRMSE$_{\text{avg}}$) for Linear Model (LM), Feed-Forward Neural Network (FFNN), and Recurrent Neural Network (RNN) prediction compared with Musculoskeletal model outputs. The FFNN and RNN models consistently outperform the linear model. For a given output category, averaging is done over all output features and test trials (with the corresponding standard deviation shown as an error bar). See Supplementary Tables \ref{tab:result1} and \ref{tab:result2} for corresponding numerical data.
        }
    \label{fig:model_comp}
    \end{figure*}

    Figure~\ref{fig:model_comp} summarises mean r and NRMSE values (along with corresponding standard deviation) for each kinematic and kinetic output category (where the mean and standard deviation are reported over all output features in a particular category) for linear, FFNN, and RNN models with numerical data reported in the Supplementary Table~\ref{tab:result1}.
    The linear model performed reasonably well for SE settings (NRMSE$_\text{avg} = 0.65_{\pm0.40}$ and r$_\text{avg} = 0.80_{\pm0.21}$); however, compared to our previous work \cite{sharma2022machine} mapping the IMC inputs (considered in this study) to the corresponding IMC-driven MSK outputs, the performance of the linear model has considerably reduced. 
    In contrast, the performance of linear models for SN settings is considerably poor, particularly in terms of NRMSE$_\text{avg} = 1.78_{\pm1.46}$, while the trends in the MSK outputs are reasonably captured with r$_\text{avg} = 0.71_{\pm0.37}$.

    Performing a rigorous hyperparameter search is essential to achieving good model fit and generalisability. In Tables~\ref{tab:NN_param} and ~\ref{tab:RNN_param}, we have listed the various hyperparameter choices (43,740 and 23,328 combinations) considered for FFNN and RNN models, respectively.
    Four-fold cross-validation was performed for each hyperparameter choice, and the model with the best validation accuracy was selected.
    The best-performing model (as listed as optimal hyperparameters in Tables~\ref{tab:NN_param} and \ref{tab:RNN_param}) was then re-trained on the training and validation data, with the final metrics reported for a held-out test dataset.
    We obtained the best performance for the FFNN architectures containing multiple hidden layers (greater than four in most cases) and wide layers containing up to 1800 nodes with the network initialised using either Xavier or Random normal initialization and parameters trained using the ReLU activation function (on hidden layers), Adam optimizer, dropout regularization, and a learning rate of 0.001 or 0.005.
    In the case of RNN models, we have treated various choices in RNN cells as hyperparameters with the observation that (B)LSTM cells perform the best for most cases with Adam or RMSprop as the optimal optimizer. 
    Interestingly, the best-fit RNN models are simpler than FFNN, with shallower and narrower networks containing only 1 or 2 RNN layers and 128 or 256 nodes in each layer.
    The same observation can be made based on the number of NN parameters reported in Supplementary Figure~\ref{fig:params}, where best-fit RNN models have fewer parameters than the best-fit FFNN models. 
    Also, the best-fit SN models contain more parameters than the SE models for the same MSK model output category.

The two NN models consistently outperformed the linear model in terms of both r$_{\text{avg}}$ and NRMSE$_{\text{avg}}$, with notable improvements in SN settings.
The performance of the FFNN and RNN models are comparable, with one model performing better for certain MSK outputs, as can be seen in Figure~\ref{fig:model_comp}.
The best-fit FFNN models have shown a high correspondence with the OMC-driven MSK outputs in both SE (r$_{\text{avg}} = 0.90_{\pm0.19}$ and NRMSE$_{\text{avg}} = 0.33_{\pm0.18}$) and SN settings (r$_{\text{avg}} = 0.84_{\pm0.23}$ and NRMSE$_{\text{avg}} = 0.85_{\pm0.79}$) with comparable statistics for RNN models with r$_{\text{avg, SE}} = 0.89_{\pm0.17}$, NRMSE$_{\text{avg, SE}} = 0.36_{\pm0.19}$, r$_{\text{avg, SN}} = 0.78_{\pm0.23}$ and NRMSE$_{\text{avg, SN}} = 0.87_{\pm0.77}$).
As can be expected, the SE models performed better than the SN models consistently across various ML model choices.

\begin{figure*}[htb]
        \centering
        \includegraphics[width=\textwidth]{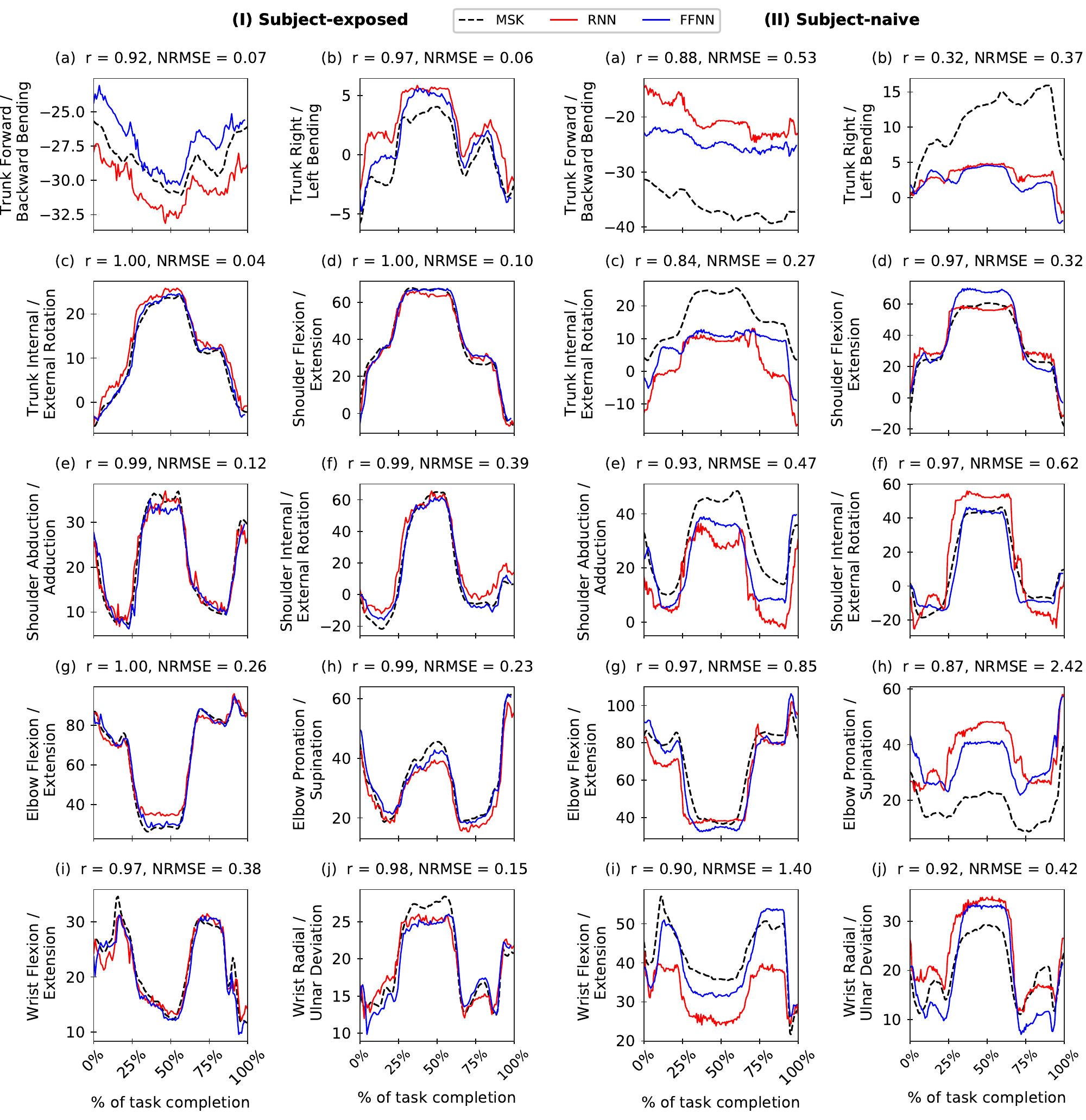}
        \caption{Comparing Feed-Forward Neural Network (FFNN) and Recurrent Neural Network (RNN) predictions for joint angles (in degrees) with the corresponding musculoskeletal (MSK) model outputs on a test~trial in \textit{Subject-exposed} (left) and \textit{Subject-naive} (right) settings. 
        Note: Performance of FFNN and RNN are comparable (see Figure \ref{fig:model_comp}); in this figure, we report r and NRMSE values only for FFNN.
        }
        \label{fig:JA}
    \end{figure*}

\begin{figure*}
        \centering    
    \includegraphics[width=\textwidth]{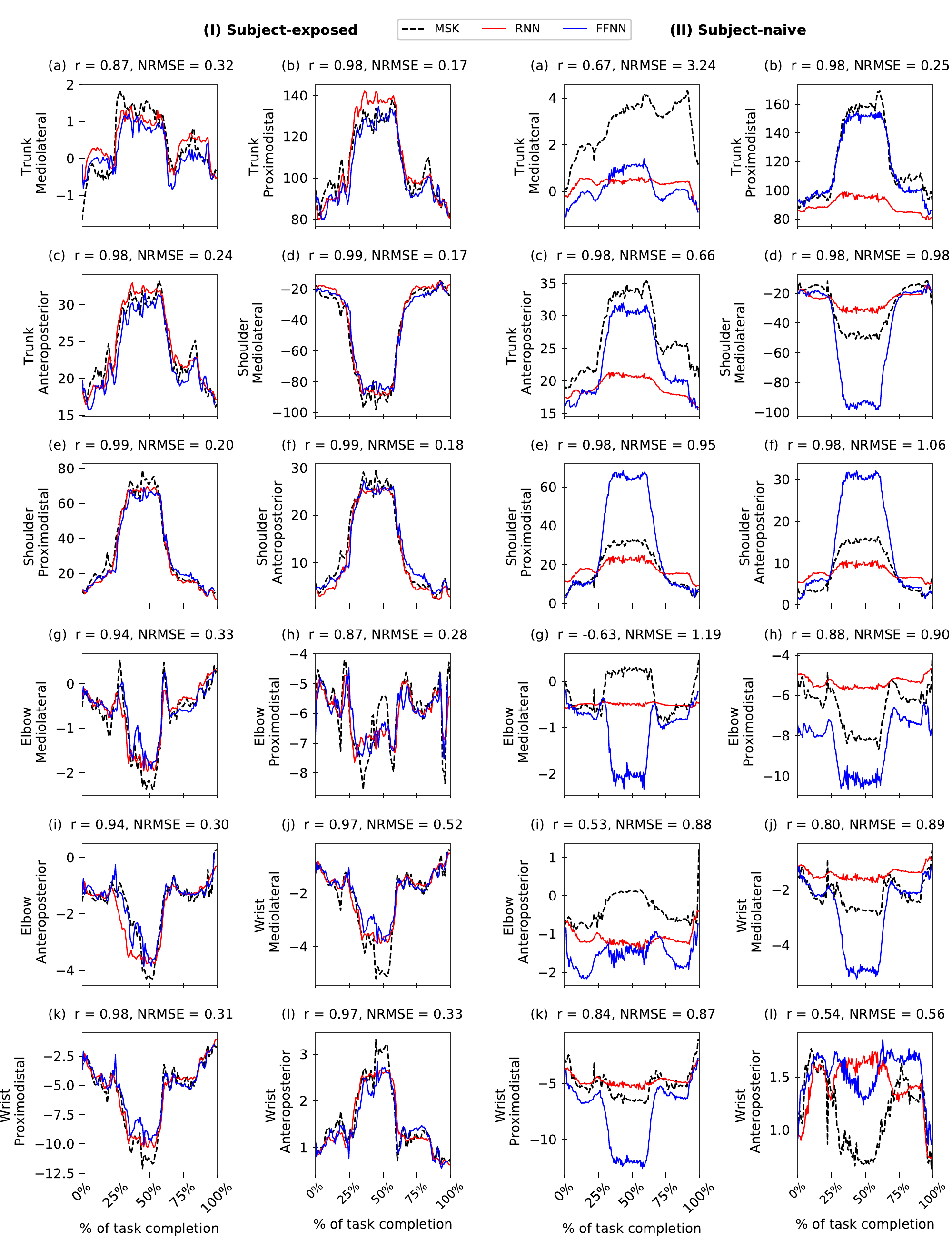}
    \caption{
        Comparing Feed-Forward Neural Network (FFNN) and Recurrent Neural Network (RNN) predictions for joint reaction forces (\% Body Weight) with  corresponding musculoskeletal (MSK) model outputs on a test~trial in \textit{Subject-exposed} (left) and \textit{Subject-naive} (right) settings. 
        Note: Performance of FFNN and RNN are comparable (see Figure \ref{fig:model_comp}); in this figure, we report r and NRMSE values only for FFNN.
        }
    \label{fig:JRF}
    \end{figure*}

\begin{figure}[htp]
        \centering
        \includegraphics[width=1\textwidth]{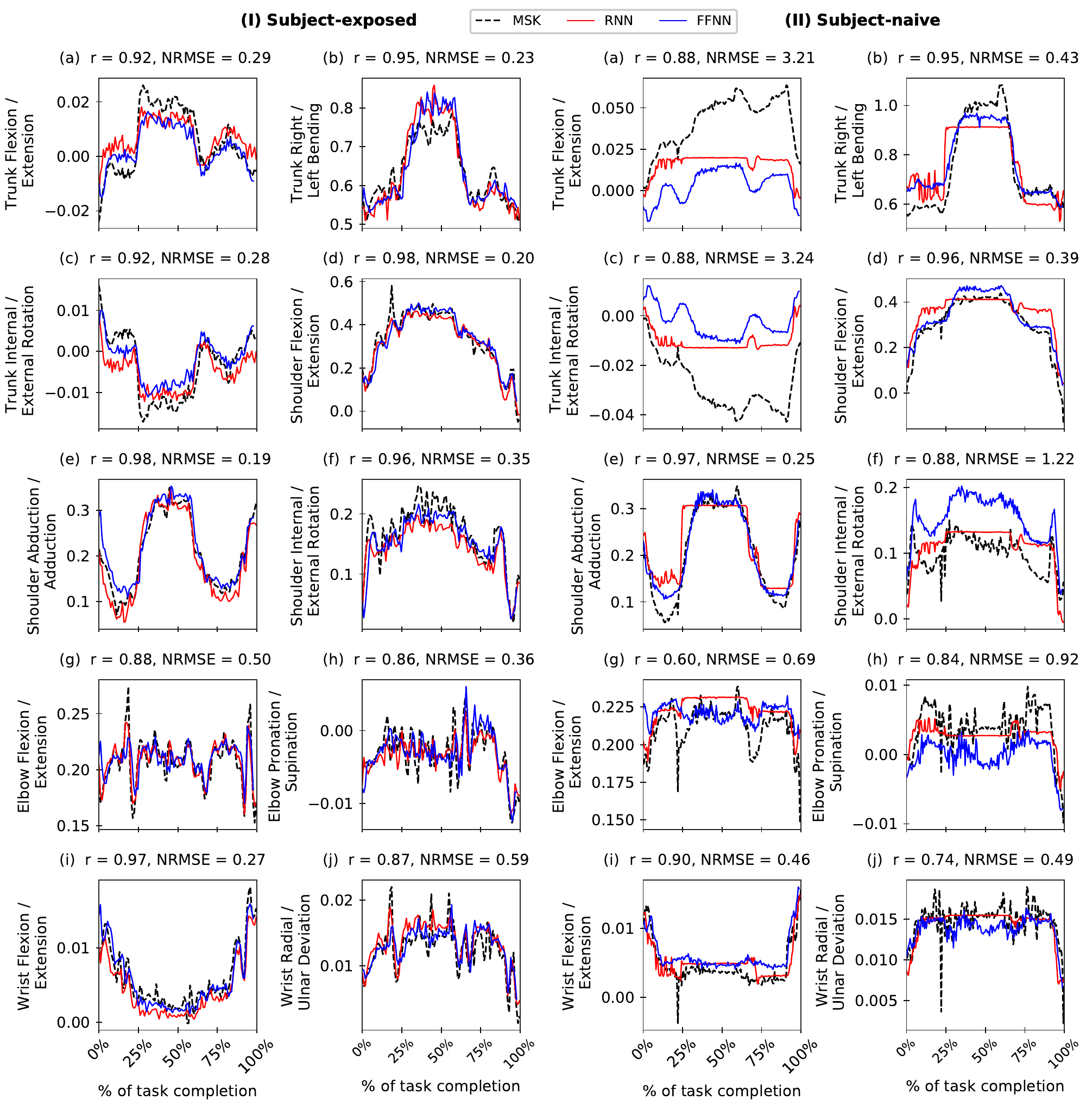}
    \caption{
        Comparing Feed-Forward Neural Network (FFNN) and Recurrent Neural Network (RNN) predictions for joint moments (\% Body Weight $\times$ Body Height) with the corresponding Musculoskeletal (MSK) model outputs on a test~trial in \textit{Subject-exposed} (left) and \textit{Subject-naive} (right) settings. Note: Performance of FFNN and RNN are comparable (see Figure 4 in the main text); in this figure, we report r and NRMSE values only for FFNN.
        }
    \label{fig:JM}
    \end{figure}

\begin{figure}[htp]
        \centering
        \includegraphics[width=1\textwidth]{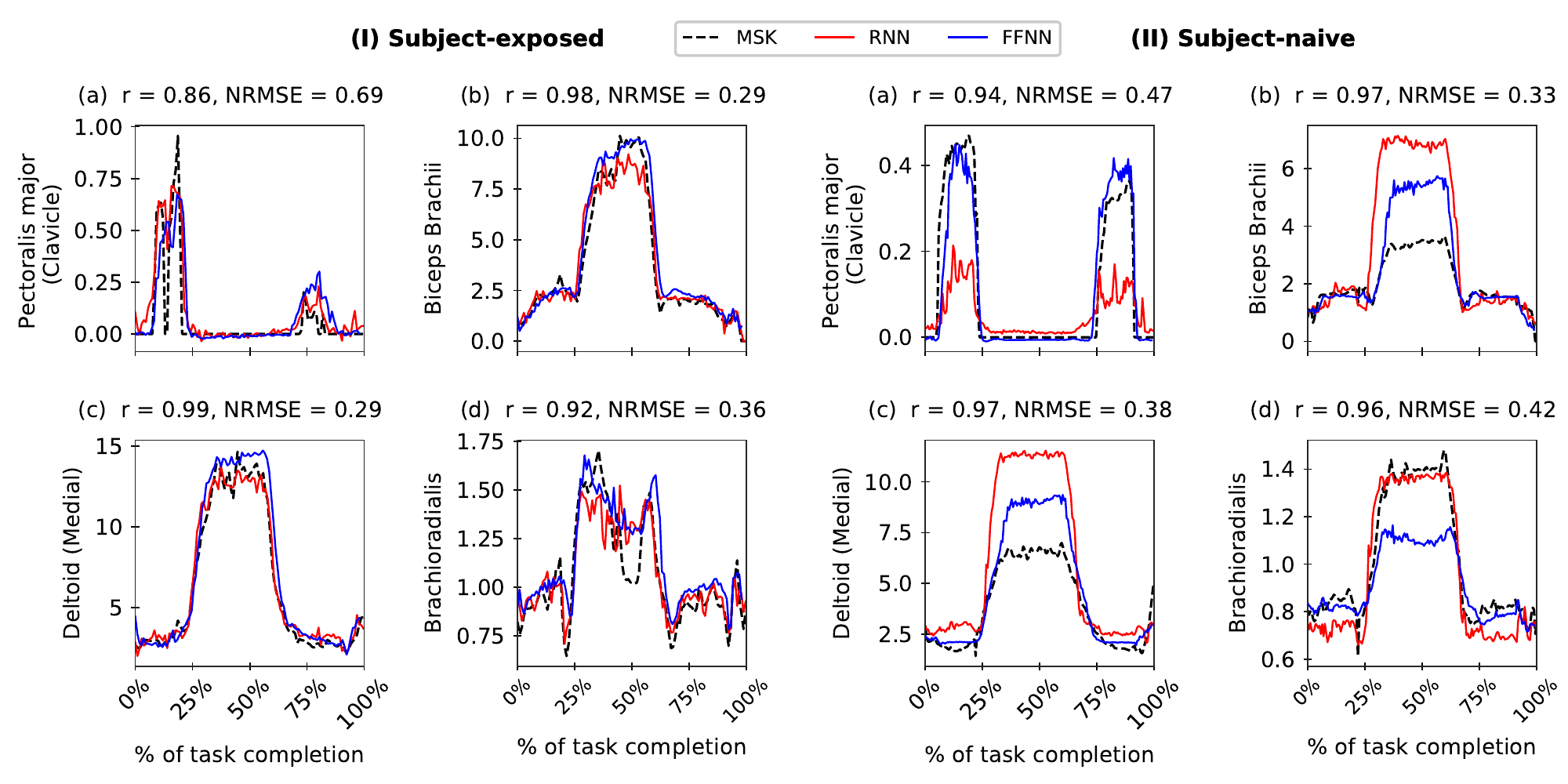}
        \caption{
        Comparing Feed-Forward Neural Network (FFNN) and Recurrent Neural Network (RNN) predictions for muscle forces (\% Body Weight) with the corresponding Musculoskeletal (MSK) model outputs for a test trial in \textit{Subject-exposed} (left) and \textit{Subject-naive} (right) settings. 
        Note: Performance of FFNN and RNN are comparable (see Figure 4 in the main text); in this figure, we report r and NRMSE values only for FFNN.
        }
    \label{fig:MF}
    \end{figure}

\begin{figure}[htp]
        \centering
        \includegraphics[width=1\textwidth]{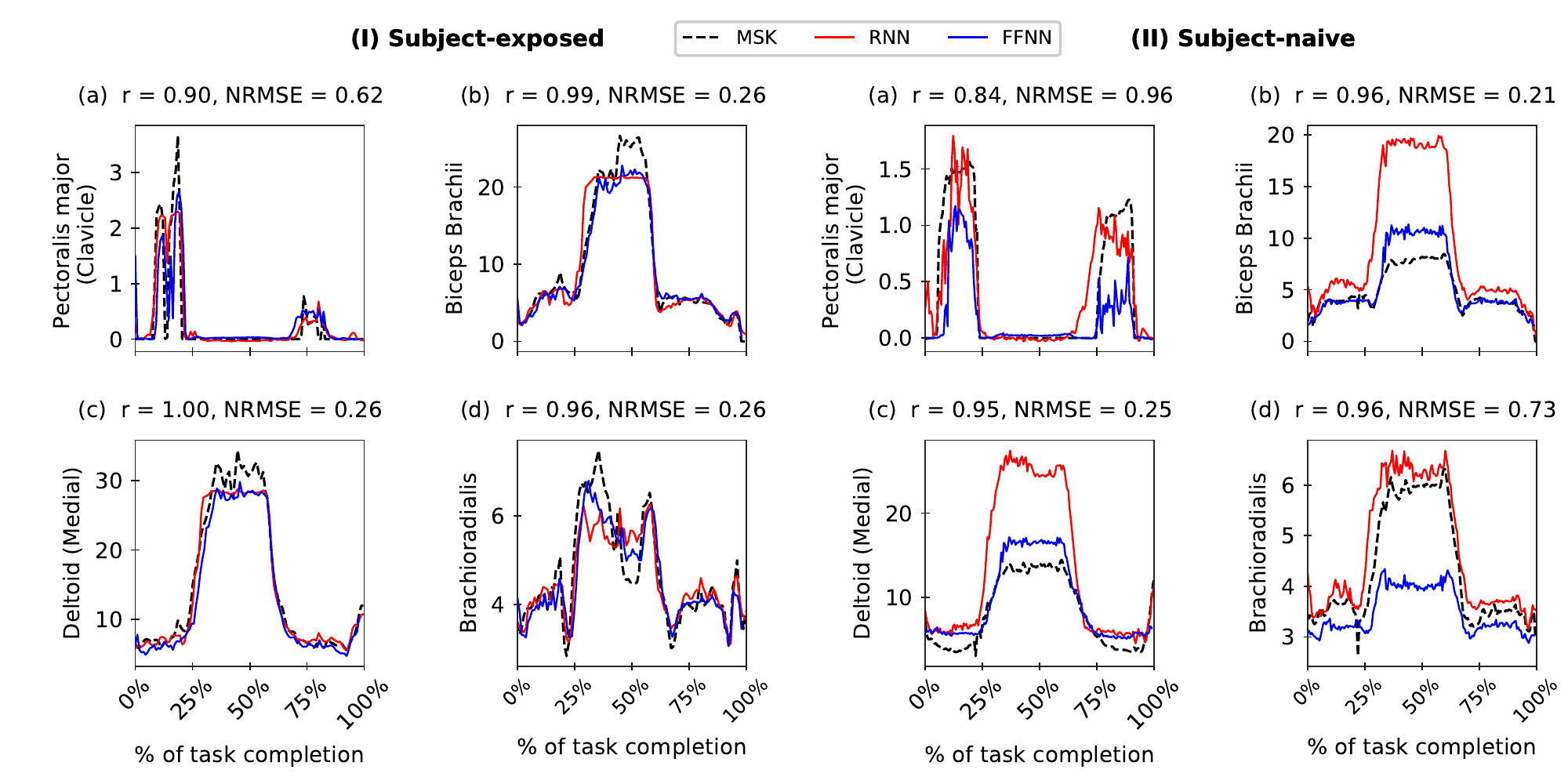}
        \caption{
        Comparing Feed-Forward Neural Network (FFNN) and Recurrent Neural Network (RNN) predictions for muscle activations (\%) with the corresponding Musculoskeletal (MSK) model outputs for a test trial in \textit{Subject-exposed} (left) and \textit{Subject-naive} (right) settings. 
        Note: Performance of FFNN and RNN are comparable (see Figure 4 in the main text); in this figure, we report r and NRMSE values only for FFNN.
        }
    \label{fig:MA}
    \end{figure}

To further aid in the interpretation of results, Figures \ref{fig:JA} and
\ref{fig:JRF} (along with Supplementary Figures~\ref{fig:JM},\ref{fig:MF},\ref{fig:MA})
show the correspondence between the NN-predicted outputs and the actual OMC-driven MSK outputs for the held-out test trial data.
Most plots show excellent r ($>$ 0.85) with notable exceptions, particularly for SN settings, in the case of Elbow Mediolateral, Elbow Anteroposterior, Wrist Anteroposterior, and Trunk Mediolateral (joint reaction forces) and Trunk Flexion/Extension and Trunk Internal/External rotation (joint moments).
For further information, we have tabulated with maximum, minimum, and inter-quartile ranges for r$_\text{avg}$ and NRMSE$_{\text{avg}}$ for all output features in Supplementary Table~\ref{tab:result2}.
Moreover, in Supplementary Table~\ref{tab:result3}, we have provided RMSE$_{\text{avg}}$ (with units) that gives an absolute error in the prediction of a given MSK output. 
Note that the interpretation of RMSE values comes with the caveat that different features cannot be compared directly due to their different scales.

\section{Discussion}\label{sec:discuss}
      
Here, ML methods are introduced to estimate OMC-driven MSK model outputs from experimental input data captured synchronously from a full-body IMC system. 
This helps bypass a computationally expensive MSK model \textit{and} obtain the corresponding OMC-driven MSK outputs (non-invasive `gold standard') while not needing the laborious and laboratory-bound OMC setup. 
The IMC system provides unprecedented ease of use, high-quality data (in near-real time), and a relatively short setup time. 
Compared to OMC systems, this method saves the time and effort to skillfully palpate bony landmarks for marker placement and cumbersome post-processing of markers or missing data.

We considered both \textit{subject-naive} (SN) and \textit{subject-exposed} (SE) scenarios -- considering both are important \cite{halilaj2018machine}, as the choice affects generalisability, and thus, model accuracy \cite{giarmatzis2020real,rane2019deep}.
While SE models tend to perform better, they
are not as generalisable to unseen subjects as SN models. SE models can incorporate subject-specific tendencies to better predict kinematic and kinetic variables corresponding to them.
Various architectures were used, ranging from a simple linear model to deep
NNs. While the linear model was adequate for SE, as expected, it could not reproduce results well in the SN case. Subsequently, we considered more complex architectures, such as
FFNNs and RNNs.
As opposed to our previous work \cite{sharma2022machine}, the discrepancies in metrics between linear models and both FFNN and RNN models are substantial, even for
SE setting, which might be attributed to the presumably greater complexity in mapping OMC-driven MSK model output to IMC input in contrast to mapping IMC-driven MSK model output to the same IMC input.
This is also corroborated by the observation that
the FFNN models implemented in this study are more complicated (Table \ref{tab:NN_param}), generally requiring more hidden layers and neurons than a similar previous study \cite{sharma2022machine}.

\begin{figure*}[htbp!]
        \centering
        \includegraphics[width=\textwidth]{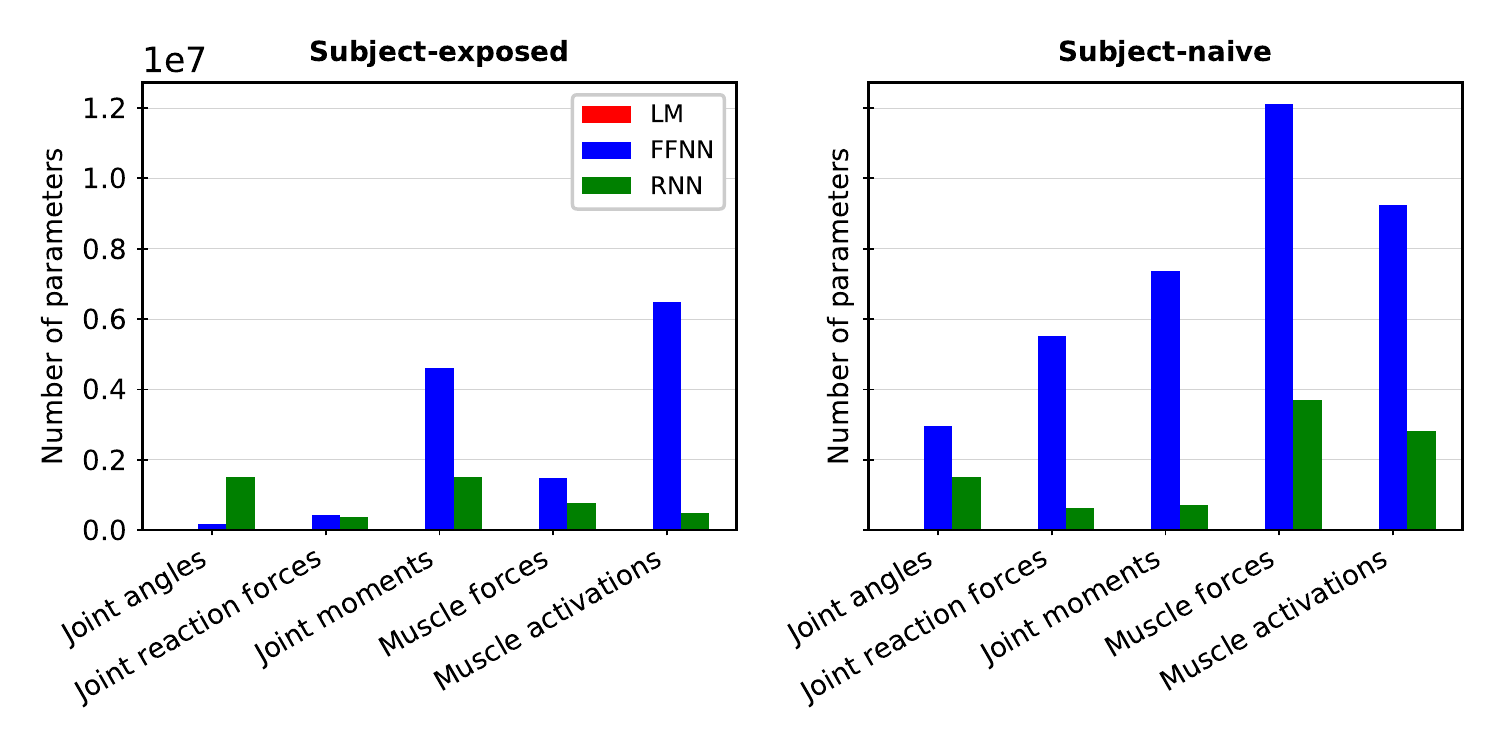}
        \caption{
            Number of parameters in Linear Model (LM), Feed-Forward Neural Network (FFNN), and Recurrent Neural Network (RNN). 
            Note: The number of parameters for Linear Models are negligible than the other two models, hence, are indiscernible here.}
    \label{fig:params}
    \end{figure*}

\begin{figure*}[htp]
        \centering
        \includegraphics[width=1\textwidth]{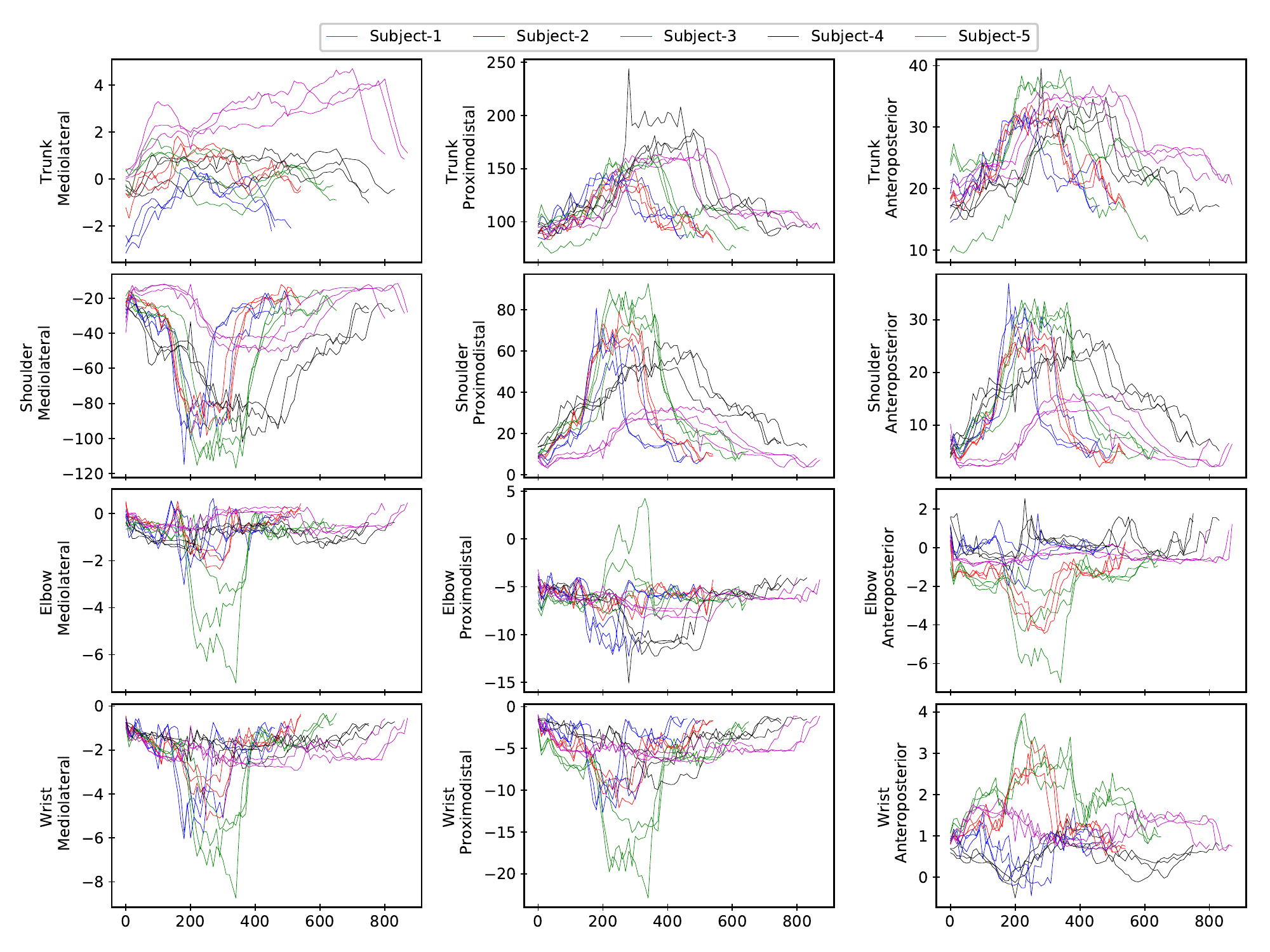}
        \caption{Joint reaction forces estimated by Musculoskeletal (MSK) model for all five subjects. Note: The three trials for individual subjects are shown in the same colour.
        }
        \label{fig:MSK_JRF_data}
    \end{figure*}

In general, we obtained excellent correspondence in both FFNN and RNN predictions with the ground truth, i.e., OMC-driven MSK output in both SE and SN scenarios (Figure \ref{fig:model_comp} and Supplementary Tables \ref{tab:result1}, \ref{tab:result2}, and \ref{tab:result3}), with some notable exceptions, in particular, for SN settings. 
Some of these issues could be attributed to the lack of adequate training data (we have $N=5$), which we hope to address in future work by adding subjects and considering synthetic data augmentation using biomechanical models \cite{mundt2020estimation,bicer2022generative}.
Note that the held-out trials were selected randomly for SE and SN settings from data with both inter- and intra-subject variability. This variability can be visualised in MSK outputs, for example, refer to Supplementary Figure~\ref{fig:MSK_JRF_data} for joint reaction forces estimated by the MSK model.
Despite this variability and small sample size ($N=5$), the FFNN and RNN models show a high degree of agreement and low NRMSE values. The predictions could be further improved with a much larger sample size.
Furthermore, the performance of FFNN and RNN models are comparable in this study despite the number of parameters in FFNN models being significantly greater than the corresponding RNN models (refer Supplementary Figure~\ref{fig:params}).
Such RNN architectures capturing the underlying MSK function using lesser model parameters can be trained easily. It would be worth exploring their performance for a much larger sample size.

Previously, FFNN was shown to predict IMC-driven MSK model outputs with high accuracy using IMC data \cite{sharma2022machine}. 
The performance of linear models and FFNNs is relatively modest in the present study than in Sharma et al. \cite{sharma2022machine}.
Notably, both OMC and IMC mocap systems are prone to soft tissue artefacts \cite{Leardini2005,Zhou2008} and instrumental errors \cite{Chiari2005} differently, which often leads to differences in pose estimation accuracy between the measurement systems (e.g., tracking long-axis rotations like shoulder internal/external rotation or elbow pronation/supination). 
In particular, IMC systems are more susceptible to soft tissue artefacts (i.e., the requirement of at least three passive markers placed on bony landmarks of a segment in an OMC system versus at least one IMU placed in the middle of the body segment in an IMC system). Soft tissue artefacts are typically task-dependent and joint-dependent and vary based on the joint's degree of freedom \cite{camomilla2017human}. Besides, the two motion tracking systems captured at different sampling frequencies might also have introduced complexity while mapping the considered IMC inputs and OMC-driven MSK outputs in the ML pipeline. However, the influence caused by these aspects needs to be scrutinised in future studies. 
Hence, the ML models' performance in this study (versus our prior study \cite{sharma2022machine}) needs to be interpreted within this context.

As a result of the experiment being conducted in a laboratory setting, good
model performance does not necessarily imply ecological validity in terms
of replication in a real-world setting. Though the experiment task is a
constrained, cyclical, goal-oriented Reach-to-Grasp task that is
ubiquitous in everyday life \cite{harthikote2019motion}, inferences from models trained on this task may not necessarily
generalise to other settings. While a study \cite{rane2019deep} involving
ML models on the lower extremity have shown generalisability across age and
body morphologies (and thus gait); this may not be the case for the upper
extremity because of a larger range of motion, higher degrees of freedom, and the three-dimensional nature of tasks
\cite{rau2000movement,Anglin2000}. The applicability of our model for
\textit{what-if} analyses involving the task with different object weights or
differently-shaped objects and other upper-extremity motions are
yet to be explored.

So far, we have focused on this study's utility in (i) bypassing the
computationally expensive MSK model, \textit{and} (ii) obtaining gold-standard OMC-driven MSK
outputs from IMC inputs. However, more recent alternative methods
such as inverse kinematics, inverse dynamics, and static optimisation can do real-time data processing, thus alleviating one of the main issues
with using MSK models, especially in field situations. While these are
valuable alternatives to the MSK model, these approaches can only obtain real-time predictions by sacrificing
accuracy and simplifying the model. For example, taking tendon
compliance into account is essential in research that relies on the muscle-tendon interaction or
stored elastic energy in tendons (e.g., in running) \cite{hicks2015my}, and helps achieve
human-like motion \cite{wang2012optimizing,miller2012evaluation}. Such
aspects are ignored in `real-time' inverse dynamics.
To achieve real-time performance,
simplified models are numerically solved up to a small number of iterations,
which decreases accuracy. Predictions improve if the models are allowed to run longer, such as in offline scenarios
\cite{van2013real,pizzolato2017real,pizzolato2017biofeedback}. Thus these
alternative methods cannot be generalised easily and encode assumptions
about certain movements that make them limited. In contrast, ML models are
flexible while offering accuracy and speed. The trade-off between speed and
accuracy in modern non-ML-based alternatives to MSK models is
an interesting question that we hope to return to in future work.

Our work relies on MSK models, and thus, it inherits some of the underlying limitations, such as not considering soft-tissue artefacts.
While MSK models are a standard method to obtain kinematics and kinetics from motion
capture, work is ongoing
\cite{Erdemir2007,lund2012validation,hicks2015my,wagner2013consistency} in
MSK model verification \& validation (V\&V). V\&V aims to
address the limitations of the MSK model in terms of accuracy concerning higher fidelity sources such as bone-pin marker measurements
\cite{Cereatti2017}, which minimises the impact of soft-tissue artefacts.
Direct measurements in general \cite{fregly2012grand,kinney2013update} and
\emph{in vivo} joint reaction forces
\cite{Bergmann2007,Westerhoff2009a,nikooyan2010validation} using
endoprosthesis has been used to validate the MSK model. Any future improvements in the MSK model from such V\&V efforts will consequently be reflected in ML models trained on MSK model outputs.
    
This work can be used as a starting point toward implementing `lab-to-field' approach by combining ML modelling with easily obtainable IMC data. This can be used for translational research that uses MSK model outputs on IMC data obtained in real-world settings such as supermarkets and manual material handling \cite{skals2021manual1,skals2021manual2,skals2021effects,larsen2020estimation,bassani2021dataset} as a training corpus for an ensemble of ML models which can be used to facilitate clinical diagnosis rapidly. In addition to IMC data, recent work on markerless systems \cite{mathis2020primer}, such as two-dimensional recordings \cite{cleather2019neural}, to obtain mocap quality data provides another source of training data while being used without specialised equipment. 

%%%%%%%%%%%%%%%%%%%%%%%%%%%%%%%%%%%%%%%%%%%%%%

\section{Study limitations}
    Our study's key limitation is the small sample size of $N=5$; nevertheless, the methodological framework introduced in this proof-of-concept study shows that it is possible to get `OMC-quality' MSK model estimations reasonably from comparatively lower-quality IMC measurements and paves the way for more elaborate studies going forward. Many studies in this field have considered up to $N=10$ subjects \cite{halilaj2018machine}. 
    Future studies should involve a larger sample size and training dataset to help with improved predictions and a more generalisable model. 
    Finally, even though our approach to mapping highly portable IMC inputs to (non-invasive gold standard) OMC input-driven MSK outputs shows immense potential, the training such ML models is limited to synchronised IMC and OMC data capture.

\section{Recommended future work}    
    We see relevant future work proceeding broadly along three lines of inquiry. Firstly, we can incorporate our ML pipelines (presented here and earlier \cite{sharma2022machine}) in relevant open-source tools \cite{cheng2019motion,cheng2019opensource} with a graphical user interface. These interfaces can assist users in estimating kinematic and kinetic variables of interest from mocap data without utilising an MSK modelling framework.
    Secondly, utilising ML models on larger datasets and potentially working with open-access mocap data. An example of such data validated with \emph{in vivo} instrumented devices is from the \textit{Grand Knee Challenge} \cite{fregly2012grand,kinney2013update} for the lower extremity. Another prominent repository is the \textit{CMU Motion Capture Dataset} \cite{Carnegie18:online} -- one of the largest publicly available mocap datasets. 
    Thirdly, we plan to expand the study to consider other ML architectures for upper-extremity biomechanical modelling, e.g., Gaussian Processes. This could be particularly beneficial in \textit{subject-naive} settings where there is room for improvement. The benefit of Gaussian Processes compared to FFNNs and RNNs is one obtains information about uncertainty in predictions automatically. 
    Undertaking methodological variations on the ML approach, such as comparing various NN frameworks or ML methods listed in \cite{halilaj2018machine,lee2022inertial,Sohane2020} and assessing their suitability for different tasks \cite{mundt2021comparison}, could be considered. Future studies should also explore ML models' robustness in the presence of artificial noise or contamination.
    
\section{Conclusion}
        
    To the authors' knowledge, this is the first-ever study (across lower and upper-extremity) that uses experimental data captured from an IMC system to directly estimate OMC-driven MSK model outputs (kinematic and kinetic) using an ML approach, bypassing the need for a (lab-based) OMC system \textit{and} a computationally expensive MSK model. This was achieved by following ML best practices, such as an extensive hyperparameter search and utilising dropout for regularization. We used FFNN and RNN architectures, alongside linear models. Both FFNN and RNN architectures achieved comparable performance on both kinematic and kinetic variables across SE and SN settings with strong-to-excellent Pearson's correlation coefficients and outperformed linear models. Our results demonstrate the effectiveness of these models, and more broadly ML methods, in approximating MSK models.

    Such ML models are viable alternatives to the MSK modelling process requiring OMC input data, which are superior to predictive models that use only IMUs.
    The computational efficiency, ease of use, and accuracy of the ML models coupled with IMC inputs make them viable as an accessible and affordable solution, especially in resource-austere settings, for taking motion analysis and MSK modelling beyond the lab.

\section*{Data availability} 
    The datasets presented here are not readily available due to ethical and privacy reasons. The Python codes used for ML implementation and data analysis here as well as the test data are available on \href{https://github.com/rahul2512/MSK_ML_beta}{https://github.com/rahul2512/MSK\_ML\_beta}. 
    
    \section*{Author Contributions}
    VHN and CM conceived the project. VHN conceptualised the biomechanical methodology and developed the anonymised dataset. CM conceptualised the ML methodology. RS, AD, CM, and VHN contributed to data processing and development of the ML models. RS developed the ML methodology and implemented the codes for ML and analysis with contributions from CM and AD. RS, AD, CM, and VHN contributed to the analysis and writing of final results. All authors contributed to drafting, reviewing, and editing. All authors have read and agreed to the final version of the manuscript.

\section*{Acknowledgment}
    This project has benefited from the dataset acquired for a previous project that was funded through the \textit{Research Councils UK (RCUK) Digital Economy Programme grant number EP/G036861/1} (Oxford Centre for Doctoral Training in Healthcare Innovation) and the \emph{Wellcome Trust Affordable Healthcare in India Award (Grant number 103383/B/13/Z)}. CM is supported by a Fellowship with the Accelerate Program for Scientific Discovery at the Computer Laboratory, University of Cambridge. We are grateful to the \emph{Swiss National Science Foundation 200020 182184} (RS) and \emph{SCITAS, EPFL} for computational resources.

% \reftitle{References}

% \bibliography{citations}

%Bibliography
\bibliographystyle{unsrt}  
\bibliography{references}  

\begin{thebibliography}{10}

\bibitem{rau2000movement}
G~Rau, C~Disselhorst-Klug, and R~Schmidt.
\newblock Movement biomechanics goes upwards: \uppercase{f}rom the leg to the
  arm.
\newblock {\em Journal of \uppercase{b}iomechanics}, 33(10):1207--1216, 2000.

\bibitem{Anglin2000}
Carolyn Anglin and UP~Wyss.
\newblock Review of arm motion analyses.
\newblock {\em Proceedings of the \uppercase{I}nstitution of
  \uppercase{M}echanical \uppercase{E}ngineers, \uppercase{P}art \uppercase{H}:
  \uppercase{J}ournal of \uppercase{E}ngineering in \uppercase{M}edicine},
  214(5):541--555, 2000.

\bibitem{Erdemir2007}
Ahmet Erdemir, Scott McLean, Walter Herzog, and Antonie~J van~den Bogert.
\newblock Model-based estimation of muscle forces exerted during movements.
\newblock {\em Clinical \uppercase{b}iomechanics}, 22(2):131--154, 2007.

\bibitem{killen2020silico}
Bryce~A Killen, Antoine Falisse, Friedl De~Groote, and Ilse Jonkers.
\newblock In silico-enhanced treatment and rehabilitation planning for patients
  with musculoskeletal disorders: \uppercase{C}an musculoskeletal modelling and
  dynamic simulations really impact current clinical practice?
\newblock {\em Applied \uppercase{S}ciences}, 10(20):7255, 2020.

\bibitem{smith2021review}
Samuel~HL Smith, Russell~J Coppack, Antonie~J van~den Bogert, Alexander~N
  Bennett, and Anthony~MJ Bull.
\newblock Review of musculoskeletal modelling in a clinical setting:
  \uppercase{C}urrent use in rehabilitation design, surgical decision making
  and healthcare interventions.
\newblock {\em Clinical \uppercase{B}iomechanics}, page 105292, 2021.

\bibitem{fregly2021conceptual}
Benjamin~J Fregly.
\newblock A conceptual blueprint for making neuromusculoskeletal models
  clinically useful.
\newblock {\em Applied \uppercase{S}ciences}, 11(5):2037, 2021.

\bibitem{Zhou2008}
Huiyu Zhou and Huosheng Hu.
\newblock Human motion tracking for rehabilitation—\uppercase{A} survey.
\newblock {\em Biomedical \uppercase{s}ignal \uppercase{p}rocessing and
  \uppercase{c}ontrol}, 3(1):1--18, 2008.

\bibitem{Cappozzo2005}
Aurelio Cappozzo, Ugo Della~Croce, Alberto Leardini, and Lorenzo Chiari.
\newblock Human movement analysis using stereophotogrammetry: \uppercase{P}art
  1: \uppercase{t}heoretical background.
\newblock {\em Gait \& \uppercase{P}osture}, 21(2):186--196, 2005.

\bibitem{jung2022applied}
Erik Jung, Cheryl Lin, Martin Contreras, and Mircea Teodorescu.
\newblock Applied machine learning on phase of gait classification and
  joint-moment regression.
\newblock {\em Biomechanics}, 2(1):44--65, 2022.

\bibitem{iosa2016wearable}
Marco Iosa, Pietro Picerno, Stefano Paolucci, and Giovanni Morone.
\newblock Wearable inertial sensors for human movement analysis.
\newblock {\em Expert \uppercase{r}eview of \uppercase{m}edical
  \uppercase{d}evices}, 13(7):641--659, 2016.

\bibitem{picerno2021wearable}
Pietro Picerno, Marco Iosa, Clive D’Souza, Maria~Grazia Benedetti, Stefano
  Paolucci, and Giovanni Morone.
\newblock Wearable inertial sensors for human movement analysis: \uppercase{a}
  five-year update.
\newblock {\em Expert \uppercase{R}eview of \uppercase{M}edical
  \uppercase{D}evices}, pages 1--16, 2021.

\bibitem{verheul2020measuring}
Jasper Verheul, Niels~J Nedergaard, Jos Vanrenterghem, and Mark~A Robinson.
\newblock Measuring biomechanical loads in team sports--\uppercase{f}rom lab to
  field.
\newblock {\em Science and \uppercase{M}edicine in \uppercase{F}ootball},
  4(3):246--252, 2020.

\bibitem{saxby2020machine}
David~J Saxby, Bryce~Adrian Killen, C~Pizzolato, CP~Carty, LE~Diamond,
  L~Modenese, J~Fernandez, G~Davico, M~Barzan, G~Lenton, et~al.
\newblock Machine learning methods to support personalized neuromusculoskeletal
  modelling.
\newblock {\em Biomechanics and \uppercase{M}odeling in
  \uppercase{M}echanobiology}, 19(4):1169--85, 2020.

\bibitem{arac2020machine}
Ahmet Arac.
\newblock Machine learning for 3d kinematic analysis of movements in
  neurorehabilitation.
\newblock {\em Current neurology and neuroscience reports}, 20(8):1--6, 2020.

\bibitem{frangoudes2022assessing}
Fotos Frangoudes, Maria Matsangidou, Eirini~C Schiza, Kleanthis Neokleous, and
  Constantinos~S Pattichis.
\newblock Assessing human motion during exercise using machine learning: A
  literature review.
\newblock {\em IEEE Access}, 10:86874--86903, 2022.

\bibitem{Damsgaard2006}
Michael Damsgaard, John Rasmussen, S{\o}ren~T{\o}rholm Christensen, Egidijus
  Surma, and Mark De~Zee.
\newblock Analysis of musculoskeletal systems in the \uppercase{A}nybody
  \uppercase{M}odeling \uppercase{S}ystem.
\newblock {\em Simulation \uppercase{M}odelling \uppercase{P}ractice and
  \uppercase{T}heory}, 14(8):1100--1111, 2006.

\bibitem{Delp2007}
Scott~L Delp, Frank~C Anderson, Allison~S Arnold, Peter Loan, Ayman Habib,
  Chand~T John, Eran Guendelman, and Darryl~G Thelen.
\newblock \uppercase{O}pen\uppercase{S}im: \uppercase{o}pen-source software to
  create and analyze dynamic simulations of movement.
\newblock {\em IEEE \uppercase{t}ransactions on \uppercase{b}iomedical
  \uppercase{e}ngineering}, 54(11):1940--1950, 2007.

\bibitem{karatsidis2019musculoskeletal}
Angelos Karatsidis, Moonki Jung, H~Martin Schepers, Giovanni Bellusci, Mark
  de~Zee, Peter~H Veltink, and Michael~Skipper Andersen.
\newblock Musculoskeletal model-based inverse dynamic analysis under ambulatory
  conditions using inertial motion capture.
\newblock {\em Medical \uppercase{e}ngineering \& \uppercase{p}hysics},
  65:68--77, 2019.

\bibitem{Konrath2019}
Jason~M Konrath, Angelos Karatsidis, H~Martin Schepers, Giovanni Bellusci, Mark
  de~Zee, and Michael~S Andersen.
\newblock Estimation of the knee adduction moment and joint contact force
  during daily living activities using inertial motion capture.
\newblock {\em Sensors}, 19(7):1681, 2019.

\bibitem{nagarajamarker}
Vikranth~H Nagaraja, Runbei Cheng, Emily (Man~Ting) Kwong, Jeroen~H. Bergmann,
  Michael~S Andersen, and Mark~S Thompson.
\newblock Marker-based vs. inertial-based motion capture:
  \uppercase{M}usculoskeletal modelling of upper extremity kinetics.
\newblock In {\em \uppercase{ISPO} \uppercase{T}rent \uppercase{I}nternational
  \uppercase{P}rosthetics \uppercase{S}ymposium (\uppercase{TIPS}) 2019}, pages
  37--38, Manchester, United Kingdom, 2019. ISPO.

\bibitem{larsen2020estimation}
Frederik~Greve Larsen, Frederik~Petri Svenningsen, Michael~Skipper Andersen,
  Mark De~Zee, and Sebastian Skals.
\newblock Estimation of spinal loading during manual materials handling using
  inertial motion capture.
\newblock {\em Annals of \uppercase{B}iomedical \uppercase{E}ngineering},
  48(2):805--821, 2020.

\bibitem{skals2021manual2}
Sebastian Skals, R{\'u}ni Bl{\'a}foss, Lars~Louis Andersen, Michael~Skipper
  Andersen, and Mark de~Zee.
\newblock Manual material handling in the supermarket sector. \uppercase{P}art
  2: \uppercase{K}nee, spine and shoulder joint reaction forces.
\newblock {\em Applied \uppercase{E}rgonomics}, 92:103345, 2021.

\bibitem{di2022inertial}
Giacomo Di~Raimondo, Benedicte Vanwanseele, Arthur Van~der Have, Jill
  Emmerzaal, Miel Willems, Bryce~Adrian Killen, and Ilse Jonkers.
\newblock Inertial sensor-to-segment calibration for accurate 3d joint angle
  calculation for use in \uppercase{O}pen\uppercase{S}im.
\newblock {\em Sensors}, 22(9):3259, 2022.

\bibitem{philp2022international}
Fraser Philp, Robert Freeman, and Caroline Stewart.
\newblock An international survey mapping practice and barriers for upper-limb
  assessments in movement analysis.
\newblock {\em Gait \& Posture}, 2022.

\bibitem{halilaj2018machine}
Eni Halilaj, Apoorva Rajagopal, Madalina Fiterau, Jennifer~L Hicks, Trevor~J
  Hastie, and Scott~L Delp.
\newblock Machine learning in human movement biomechanics: \uppercase{B}est
  practices, common pitfalls, and new opportunities.
\newblock {\em Journal of \uppercase{b}iomechanics}, 81:1--11, 2018.

\bibitem{xiang2022recent}
Liangliang Xiang, Alan Wang, Yaodong Gu, Liang Zhao, Vickie Shim, and Justin
  Fernandez.
\newblock Recent machine learning progress in lower limb running biomechanics
  with wearable technology: A systematic review.
\newblock {\em Frontiers in Neurorobotics}, 16, 2022.

\bibitem{amrein2023machine}
Sabrina Amrein, Charlotte Werner, Ursina Arnet, and Wiebe~HK de~Vries.
\newblock Machine-learning-based methodology for estimation of shoulder load in
  wheelchair-related activities using wearables.
\newblock {\em Sensors}, 23(3):1577, 2023.

\bibitem{lee2022inertial}
Chang~June Lee and Jung~Keun Lee.
\newblock Inertial motion capture-based wearable systems for estimation of
  joint kinetics: A systematic review.
\newblock {\em Sensors}, 22(7):2507, 2022.

\bibitem{sharma2022machine}
Rahul Sharma, Abhishek Dasgupta, Runbei Cheng, Challenger Mishra, and
  Vikranth~H. Nagaraja.
\newblock Machine learning for musculoskeletal modeling of upper extremity.
\newblock {\em IEEE Sensors Journal}, 22(19), 2022.

\bibitem{wouda2018estimation}
Frank~J Wouda, Matteo Giuberti, Giovanni Bellusci, Erik Maartens, Jasper
  Reenalda, Bert-Jan~F Van~Beijnum, and Peter~H Veltink.
\newblock Estimation of vertical ground reaction forces and sagittal knee
  kinematics during running using three inertial sensors.
\newblock {\em Frontiers in \uppercase{p}hysiology}, 9:218, 2018.

\bibitem{fernandez2020population}
Justin Fernandez, Alex Dickinson, and Peter Hunter.
\newblock Population based approaches to computational musculoskeletal
  modelling, 2020.

\bibitem{Sohane2020}
Anurag Sohane and Ravinder Agarwal.
\newblock Knee muscle force estimating model using machine learning approach.
\newblock {\em The \uppercase{C}omputer \uppercase{J}ournal}, 2020.

\bibitem{mubarrat2022shoulder}
S.~T. Mubarrat and S.~Chowdhury.
\newblock Convolutional lstm: a deep learning approach to predict shoulder
  joint reaction forces.
\newblock {\em Computer Methods in Biomechanics and Biomedical Engineering},
  0(0):1--13, 2022.

\bibitem{cleather2019neural}
Daniel~J Cleather.
\newblock Neural network based approximation of muscle and joint contact forces
  during jumping and landing.
\newblock {\em Journal of \uppercase{H}uman \uppercase{P}erformance and
  \uppercase{H}ealth}, 1:f1--f13, 2019.

\bibitem{giarmatzis2020real}
Georgios Giarmatzis, Evangelia~I Zacharaki, and Konstantinos Moustakas.
\newblock Real-time prediction of joint forces by motion capture and machine
  learning.
\newblock {\em Sensors}, 20(23):6933, 2020.

\bibitem{mundt2020estimation}
Marion Mundt, Arnd Koeppe, Sina David, Tom Witter, Franz Bamer, Wolfgang
  Potthast, and Bernd Markert.
\newblock Estimation of gait mechanics based on simulated and measured
  \uppercase{IMU} data using an artificial neural network.
\newblock {\em Frontiers in \uppercase{b}ioengineering and
  \uppercase{b}iotechnology}, 8, 2020.

\bibitem{mundt2020prediction}
Marion Mundt, Arnd Koeppe, Sina David, Franz Bamer, Wolfgang Potthast, and
  Bernd Markert.
\newblock Prediction of ground reaction force and joint moments based on
  optical motion capture data during gait.
\newblock {\em Medical Engineering \& Physics}, 86:29--34, 2020.

\bibitem{mundt2020predictionimcomc}
Marion Mundt, Wolf Thomsen, Tom Witter, Arnd Koeppe, Sina David, Franz Bamer,
  Wolfgang Potthast, and Bernd Markert.
\newblock Prediction of lower limb joint angles and moments during gait using
  artificial neural networks.
\newblock {\em Medical \& biological engineering \& computing}, 58(1):211--225,
  2020.

\bibitem{nagaraja2018compensatory}
Vikranth Nagaraja, Jeroen Bergmann, Michael~Skipper Andersen, and Mark
  Thompson.
\newblock Compensatory movements involved during simulated upper limb
  prosthetic usage: \uppercase{R}each \uppercase{T}ask vs.
  \uppercase{R}each-to-\uppercase{G}rasp \uppercase{T}ask.
\newblock In {\em \uppercase{XV ISB I}nternational \uppercase{S}ymposium on
  \uppercase{3-D A}nalysis of \uppercase{H}uman \uppercase{M}ovement}, Salford,
  United Kingdom, 2018. ISB.

\bibitem{nagaraja2023comparison}
Vikranth~H Nagaraja, Jeroen~HM Bergmann, Michael~S Andersen, and Mark~S
  Thompson.
\newblock Comparison of a scaled cadaver-based musculoskeletal model with a
  clinical upper extremity model.
\newblock {\em Journal of Biomechanical Engineering}, 145(4):041012, 2023.

\bibitem{PluginGa94:online}
Vicon motion systems $|$ \uppercase{P}lug-in \uppercase{G}ait
  \uppercase{R}eference \uppercase{G}uide -- \uppercase{V}icon
  \uppercase{D}ocumentation.
\newblock
  \url{https://docs.vicon.com/display/Nexus212/Plug-in+Gait+Reference+Guide}.
\newblock (Accessed on 06/02/2023).

\bibitem{paulich2018xsens}
Monique Paulich, Martin Schepers, Nina Rudigkeit, and Giovanni Bellusci.
\newblock Xsens mtw awinda: \uppercase{M}iniature wireless inertial-magnetic
  motion tracker for highly accurate 3\uppercase{D} kinematic applications.
\newblock {\em Xsens: \uppercase{E}nschede, \uppercase{T}he
  \uppercase{N}etherlands}, pages 1--9, 2018.

\bibitem{MVNUserM34:online}
Xsens \uppercase{MVN U}ser \uppercase{M}anual.
\newblock
  \url{https://www.xsens.com/hubfs/Downloads/usermanual/MVN_User_Manual.pdf},
  2021.
\newblock (Accessed on 06/02/2023).

\bibitem{Synchron68:online}
Xsens.
\newblock Syncronising \uppercase{X}sens systems with \uppercase{V}icon
  \uppercase{N}exus.
\newblock
  \url{https://www.xsens.com/hubfs/Downloads/plugins\%20\%20tools/SynchronisingXsenswithVicon.pdf},
  January 2020.
\newblock (Accessed on 06/02/2023).

\bibitem{PDFdownl80:online}
Vicon nexus™ 2.5 manual. 2016. what’s new in vicon nexus 2.5 – nexus 2.5
  documentation. documentation docs.vicon.com.
\newblock
  \url{https://docs.vicon.com/display/Nexus25/PDF+downloads+for+Vicon+Nexus?preview=/50888706/50889382/ViconNexusWhatsNew25.pdf}.
\newblock (Accessed on 06/02/2023).

\bibitem{C3DORGTh92:online}
{Motion Lab Systems}.
\newblock \uppercase{C3D.ORG} - the biomechanics standard file format.
\newblock \url{https://www.c3d.org/}.
\newblock (Accessed on 06/02/2023).

\bibitem{MVNAnaly30:online}
\uppercase{MVN A}nalyze.
\newblock
  \url{https://www.movella.com/products/motion-capture/mvn-analyze#overview}.
\newblock (Accessed on 06/02/2023).

\bibitem{harthikote2019motion}
Vikranth Harthikote~Nagaraja.
\newblock {\em Motion capture and musculoskeletal simulation tools to measure
  prosthetic arm functionality}.
\newblock PhD thesis, Department of Engineering Science, University of Oxford,
  Oxford, United Kingdom, 2019.

\bibitem{moisio2003normalization}
Kirsten~C Moisio, Dale~R Sumner, Susan Shott, and Debra~E Hurwitz.
\newblock Normalization of joint moments during gait: \uppercase{a} comparison
  of two techniques.
\newblock {\em Journal of \uppercase{b}iomechanics}, 36(4):599--603, 2003.

\bibitem{derrick2020isb}
Timothy~R Derrick, Antonie~J van~den Bogert, Andrea Cereatti, Rapha{\"e}l
  Dumas, Silvia Fantozzi, and Alberto Leardini.
\newblock \uppercase{ISB} recommendations on the reporting of intersegmental
  forces and moments during human motion analysis.
\newblock {\em Journal of \uppercase{b}iomechanics}, 99:109533, 2020.

\bibitem{Bergmann2007}
G~Bergmann, F~Graichen, A~Bender, M~K{\"a}{\"a}b, A~Rohlmann, and P~Westerhoff.
\newblock In vivo glenohumeral contact forces—measurements in the first
  patient 7 months postoperatively.
\newblock {\em Journal of \uppercase{b}iomechanics}, 40(10):2139--2149, 2007.

\bibitem{Goodfellow-et-al-2016}
Ian Goodfellow, Yoshua Bengio, and Aaron Courville.
\newblock {\em Deep Learning}.
\newblock MIT Press, 2016.
\newblock \url{http://www.deeplearningbook.org}.

\bibitem{hochreiter1997long}
Sepp Hochreiter and J{\"u}rgen Schmidhuber.
\newblock Long short-term memory.
\newblock {\em Neural computation}, 9(8):1735--1780, 1997.

\bibitem{cho2014properties}
Kyunghyun Cho, Bart Van~Merri{\"e}nboer, Dzmitry Bahdanau, and Yoshua Bengio.
\newblock On the properties of neural machine translation: Encoder-decoder
  approaches.
\newblock {\em arXiv preprint arXiv:1409.1259}, 2014.

\bibitem{rane2019deep}
Lance Rane, Ziyun Ding, Alison~H McGregor, and Anthony~MJ Bull.
\newblock Deep learning for musculoskeletal force prediction.
\newblock {\em Annals of \uppercase{b}iomedical \uppercase{e}ngineering},
  47(3):778--789, 2019.

\bibitem{bicer2022generative}
Metin Bicer, Andrew~TM Phillips, Alessandro Melis, Alison~H McGregor, and Luca
  Modenese.
\newblock Generative deep learning applied to biomechanics: A new augmentation
  technique for motion capture datasets.
\newblock {\em Journal of Biomechanics}, page 111301, 2022.

\bibitem{Leardini2005}
Alberto Leardini, Lorenzo Chiari, Ugo Della~Croce, and Aurelio Cappozzo.
\newblock Human movement analysis using stereophotogrammetry: \uppercase{P}art
  3. \uppercase{S}oft tissue artifact assessment and compensation.
\newblock {\em Gait \& posture}, 21(2):212--225, 2005.

\bibitem{Chiari2005}
Lorenzo Chiari, Ugo Della~Croce, Alberto Leardini, and Aurelio Cappozzo.
\newblock Human movement analysis using stereophotogrammetry: \uppercase{P}art
  2: \uppercase{I}nstrumental errors.
\newblock {\em Gait \& posture}, 21(2):197--211, 2005.

\bibitem{camomilla2017human}
Valentina Camomilla, Rapha{\"e}l Dumas, and Aurelio Cappozzo.
\newblock Human movement analysis: \uppercase{T}he soft tissue artefact issue.
\newblock {\em Journal of \uppercase{b}iomechanics}, 62:pp--1, 2017.

\bibitem{hicks2015my}
Jennifer~L Hicks, Thomas~K Uchida, Ajay Seth, Apoorva Rajagopal, and Scott~L
  Delp.
\newblock Is my model good enough? \uppercase{B}est practices for verification
  and validation of musculoskeletal models and simulations of movement.
\newblock {\em Journal of \uppercase{b}iomechanical \uppercase{e}ngineering},
  137(2), 2015.

\bibitem{wang2012optimizing}
Jack~M Wang, Samuel~R Hamner, Scott~L Delp, and Vladlen Koltun.
\newblock Optimizing locomotion controllers using biologically-based actuators
  and objectives.
\newblock {\em ACM \uppercase{T}ransactions on \uppercase{G}raphics
  (\uppercase{TOG})}, 31(4):1--11, 2012.

\bibitem{miller2012evaluation}
Ross~H Miller, Brian~R Umberger, Joseph Hamill, and Graham~E Caldwell.
\newblock Evaluation of the minimum energy hypothesis and other potential
  optimality criteria for human running.
\newblock {\em Proceedings of the \uppercase{R}oyal \uppercase{S}ociety
  \uppercase{B}: \uppercase{B}iological \uppercase{S}ciences},
  279(1733):1498--1505, 2012.

\bibitem{van2013real}
Antonie~J Van~den Bogert, Thomas Geijtenbeek, Oshri Even-Zohar, Frans
  Steenbrink, and Elizabeth~C Hardin.
\newblock A real-time system for biomechanical analysis of human movement and
  muscle function.
\newblock {\em Medical \& \uppercase{b}iological \uppercase{e}ngineering \&
  \uppercase{c}omputing}, 51(10):1069--1077, 2013.

\bibitem{pizzolato2017real}
C~Pizzolato, M~Reggiani, L~Modenese, and DG~Lloyd.
\newblock Real-time inverse kinematics and inverse dynamics for lower limb
  applications using \uppercase{O}pen\uppercase{S}im.
\newblock {\em Computer \uppercase{m}ethods in \uppercase{b}iomechanics and
  \uppercase{b}iomedical \uppercase{e}ngineering}, 20(4):436--445, 2017.

\bibitem{pizzolato2017biofeedback}
Claudio Pizzolato, Monica Reggiani, David~J Saxby, Elena Ceseracciu, Luca
  Modenese, and David~G Lloyd.
\newblock Biofeedback for gait retraining based on real-time estimation of
  tibiofemoral joint contact forces.
\newblock {\em IEEE \uppercase{T}ransactions on \uppercase{N}eural
  \uppercase{S}ystems and \uppercase{R}ehabilitation \uppercase{E}ngineering},
  25(9):1612--1621, 2017.

\bibitem{lund2012validation}
Morten~Enemark Lund, Mark de~Zee, Michael~Skipper Andersen, and John Rasmussen.
\newblock On validation of multibody musculoskeletal models.
\newblock {\em Proceedings of the \uppercase{I}nstitution of
  \uppercase{M}echanical \uppercase{E}ngineers, \uppercase{P}art \uppercase{H}:
  \uppercase{J}ournal of \uppercase{E}ngineering in \uppercase{M}edicine},
  226(2):82--94, 2012.

\bibitem{wagner2013consistency}
David~W Wagner, Vahagn Stepanyan, James~M Shippen, Matthew~S DeMers, Robin~S
  Gibbons, Brian~J Andrews, Graham~H Creasey, and Gary~S Beaupre.
\newblock Consistency among musculoskeletal models: \uppercase{c}aveat
  utilitor.
\newblock {\em Annals of \uppercase{b}iomedical \uppercase{e}ngineering},
  41(8):1787--1799, 2013.

\bibitem{Cereatti2017}
Andrea Cereatti, Tecla Bonci, Massoud Akbarshahi, Kamiar Aminian, Arnaud
  Barr{\'e}, Mickael Begon, Daniel~L Benoit, Caecilia Charbonnier, Fabien
  Dal~Maso, et~al.
\newblock Standardization proposal of soft tissue artefact description for data
  sharing in human motion measurements.
\newblock {\em Journal of biomechanics}, 62:5--13, 2017.

\bibitem{fregly2012grand}
Benjamin~J Fregly, Thor~F Besier, David~G Lloyd, Scott~L Delp, Scott~A Banks,
  Marcus~G Pandy, and Darryl~D D'lima.
\newblock Grand challenge competition to predict in vivo knee loads.
\newblock {\em Journal of \uppercase{o}rthopaedic \uppercase{r}esearch},
  30(4):503--513, 2012.

\bibitem{kinney2013update}
Allison~L Kinney, Thor~F Besier, Darryl~D D'Lima, and Benjamin~J Fregly.
\newblock Update on grand challenge competition to predict in vivo knee loads.
\newblock {\em Journal of \uppercase{b}iomechanical \uppercase{e}ngineering},
  135(2), 2013.

\bibitem{Westerhoff2009a}
P~Westerhoff, F~Graichen, A~Bender, A~Halder, A~Beier, A~Rohlmann, and
  G~Bergmann.
\newblock In vivo measurement of shoulder joint loads during activities of
  daily living.
\newblock {\em Journal of \uppercase{b}iomechanics}, 42(12):1840--1849, 2009.

\bibitem{nikooyan2010validation}
A~Asadi Nikooyan, HEJ Veeger, P~Westerhoff, F~Graichen, G~Bergmann, and FCT
  Van~der Helm.
\newblock Validation of the \uppercase{D}elft \uppercase{S}houlder and
  \uppercase{E}lbow \uppercase{M}odel using in-vivo glenohumeral joint contact
  forces.
\newblock {\em Journal of \uppercase{b}iomechanics}, 43(15):3007--3014, 2010.

\bibitem{skals2021manual1}
Sebastian Skals, R{\'u}ni Bl{\'a}foss, Michael~Skipper Andersen, Mark de~Zee,
  and Lars~Louis Andersen.
\newblock Manual material handling in the supermarket sector. \uppercase{P}art
  1: \uppercase{J}oint angles and muscle activity of trapezius descendens and
  erector spinae longissimus.
\newblock {\em Applied \uppercase{E}rgonomics}, 92:103340, 2021.

\bibitem{skals2021effects}
Sebastian Skals, R{\'u}ni Bl{\'a}foss, Mark de~Zee, Lars~Louis Andersen, and
  Michael~Skipper Andersen.
\newblock Effects of load mass and position on the dynamic loading of the
  knees, shoulders and lumbar spine during lifting: \uppercase{a}
  musculoskeletal modelling approach.
\newblock {\em Applied \uppercase{E}rgonomics}, 96:103491, 2021.

\bibitem{bassani2021dataset}
Giulia Bassani, Alessandro Filippeschi, and Carlo~Alberto Avizzano.
\newblock A dataset of human motion and muscular activities in manual material
  handling tasks for biomechanical and ergonomic analyses.
\newblock {\em IEEE \uppercase{S}ensors \uppercase{J}ournal},
  21(21):24731--24739, 2021.

\bibitem{mathis2020primer}
Alexander Mathis, Steffen Schneider, Jessy Lauer, and Mackenzie~Weygandt
  Mathis.
\newblock A primer on motion capture with deep learning:
  \uppercase{p}rinciples, pitfalls, and perspectives.
\newblock {\em Neuron}, 108(1):44--65, 2020.

\bibitem{cheng2019motion}
Runbei Cheng, Vikranth~H. Nagaraja, Jeroen~H. Bergmann, and Mark~S. Thompson.
\newblock \uppercase{M}otion \uppercase{C}apture \uppercase{A}nalysis \&
  \uppercase{P}lotting \uppercase{A}ssistant: \uppercase{A}n opensource
  framework to analyse inertial-sensor-based measurements.
\newblock In {\em \uppercase{ISPO T}rent \uppercase{I}nternational
  \uppercase{P}rosthetics \uppercase{S}ymposium (\uppercase{TIPS}) 2019}, pages
  156--157, Manchester, United Kingdom, 2019. ISPO.

\bibitem{cheng2019opensource}
Runbei Cheng, Vikranth~H. Nagaraja, Jeroen~H. Bergmann, and Mark~S. Thompson.
\newblock An opensource framework to analyse marker-based and
  inertial-sensor-based measurements: \uppercase{M}otion \uppercase{C}apture
  \uppercase{A}nalysis \& \uppercase{P}lotting \uppercase{A}ssistant
  (\uppercase{MCAPA}) 2.0.
\newblock In {\em BioMedEng 2019}, London, United Kingdom, 2019.

\bibitem{Carnegie18:online}
Carnegie \uppercase{M}ellon \uppercase{U}niversity -- \uppercase{CMU}
  \uppercase{G}raphics \uppercase{L}ab - \uppercase{m}otion \uppercase{c}apture
  \uppercase{l}ibrary.
\newblock \url{http://mocap.cs.cmu.edu/}, 2003.
\newblock (Accessed on 06/02/2023).

\bibitem{mundt2021comparison}
Marion Mundt, William~R Johnson, Wolfgang Potthast, Bernd Markert, Ajmal Mian,
  and Jacqueline Alderson.
\newblock A comparison of three neural network approaches for estimating joint
  angles and moments from \uppercase{I}nertial \uppercase{M}easurement
  \uppercase{U}nits.
\newblock {\em Sensors}, 21(13):4535, 2021.

\end{thebibliography}
\pagebreak

\centering{\section*{Supplementary Information}}

\subsection*{2. Materials and methods}
\subsubsection*{2.1 Data}

\subsection*{3. Results}

\begin{table}[htp]
\begin{center}
\begin{tabular}{ c | c c c c c c }
\hline
& \multicolumn{2}{c}{\textbf{Linear Model}} & \multicolumn{2}{c}{\textbf{Feed-Forward Neural Network}} & \multicolumn{2}{c}{\textbf{Recurrent Neural Network}}\\ 
\hline
& \multicolumn{6}{c}{} \\[-0.8em]
& r$_{\text{avg}}$ & NRMSE$_{\text{avg}}$ & r$_{\text{avg}}$ & NRMSE$_{\text{avg}}$ & r$_{\text{avg}}$ & NRMSE$_{\text{avg}}$ \\
Output & Mean (SD) & Mean (SD)& Mean (SD) & Mean (SD)& Mean (SD) & Mean (SD) \\
\hline
\hline
&  \multicolumn{6}{c}{} \\[-0.8em]
 &  \multicolumn{6}{c}{\textbf{Subject-exposed}}  \\ 
\arrayrulecolor{lightgray} \hline \arrayrulecolor{black}
&  \multicolumn{6}{c}{} \\[-0.8em]
Joint angles  & 
0.85 (0.15)  &  0.64 (0.62)
 & 
0.92 (0.23)  &  0.25 (0.2)
 & 
0.93 (0.13)  &  0.32 (0.24)
\\
Joint reaction forces  & 
0.77 (0.21)  &  0.70 (0.31)
 & 
0.91 (0.12)  &  0.34 (0.11)
 & 
0.91 (0.11)  &  0.33 (0.09)
\\
Joint moments  & 
0.81 (0.11)  &  0.58 (0.25)
 & 
0.88 (0.08)  &  0.41 (0.2)
 & 
0.87 (0.09)  &  0.46 (0.2)
\\
Muscle forces  & 
0.73 (0.31)  &  0.64 (0.27)
 & 
0.85 (0.29)  &  0.37 (0.17)
 & 
0.83 (0.3)  &  0.33 (0.17)
\\
Muscle activations  & 
0.75 (0.28)  &  0.74 (0.28)
 & 
0.88 (0.25)  &  0.29 (0.15)
 & 
0.84 (0.28)  &  0.37 (0.17)
\\
\arrayrulecolor{lightgray} \hline \arrayrulecolor{black}
&&&&&& \\[-0.8em]
 &  \multicolumn{6}{c}{\textbf{Subject-naive}}  \\ 
\arrayrulecolor{lightgray} \hline \arrayrulecolor{black}
&&&&&& \\[-0.8em]
Joint angles  & 
0.85 (0.11)  &  2.54 (2.22)
 & 
0.86 (0.17)  &  0.82 (0.65)
 & 
0.86 (0.13)  &  1.10 (1.02)
\\
Joint reaction forces  & 
0.5 (0.59)  &  1.34 (0.98)
 & 
0.75 (0.35)  &  0.90 (0.74)
 & 
0.64 (0.34)  &  0.84 (0.77)
\\
Joint moments  & 
0.84 (0.11)  &  1.57 (0.68)
 & 
0.85 (0.12)  &  1.12 (1.02)
 & 
0.75 (0.13)  &  0.89 (0.68)
\\
Muscle forces  & 
0.71 (0.2)  &  1.70 (1.20)
 & 
0.95 (0.03)  &  0.49 (0.53)
 & 
0.93 (0.04)  &  0.58 (0.27)
\\
Muscle activations  & 
0.70 (0.17)  &  1.78 (1.16)
 & 
0.93 (0.04)  &  0.42 (0.28)
 & 
0.92 (0.06)  &  0.61 (0.21)
\\
\hline 
\end{tabular}
\end{center}

\caption{
Average Pearson's correlation coefficient and average NRMSE values for Linear Model (LM), Feed-Forward Neural Network (FFNN), and Recurrent Neural Network (RNN) prediction compared with Musculoskeletal (MSK) model outputs. The FFNN and RNN models consistently outperform the linear model. For a given output category, averaging is done over all output features and test trials.
}
\label{tab:result1}
\end{table}

\begin{table}[htb!]
\begin{center}
\begin{tabular}{ c | c | c c c c c | c c c c c }
 \hline
& & \multicolumn{5}{c}{r} & \multicolumn{5}{c}{NRMSE} \\ 
 \hline
 Output & ML model &  Mean & SD & Max & Min & IQR & Mean & SD  & Max & Min & IQR \\
\hline
\hline
&  & \multicolumn{10}{c}{} \\[-0.8em]
&  & \multicolumn{10}{c}{\textbf{Subject-exposed}}\\ 
&  & \multicolumn{10}{c}{} \\[-0.8em]
\arrayrulecolor{lightgray} \hline \arrayrulecolor{black}
\multirow{2}*{ Joint angles }
 & FFNN  &  0.92 & 0.23 & 1.0 & -0.07 & 0.04  &  0.25 & 0.2 & 0.79 & 0.04 & 0.28 \\
 &  BLSTM  &  0.93 & 0.13 & 0.99 & 0.38 & 0.07  &  0.32 & 0.24 & 0.76 & 0.03 & 0.45 \\
\arrayrulecolor{lightgray} \hline \arrayrulecolor{black}
\multirow{2}*{ Joint reaction forces }
 & FFNN  &  0.91 & 0.12 & 0.99 & 0.46 & 0.07  &  0.34 & 0.11 & 0.55 & 0.17 & 0.13 \\
 &  LSTM  &  0.91 & 0.11 & 0.99 & 0.51 & 0.09  &  0.33 & 0.09 & 0.49 & 0.18 & 0.16 \\
\arrayrulecolor{lightgray} \hline \arrayrulecolor{black}
\multirow{2}*{ Joint moments }
 & FFNN  &  0.88 & 0.08 & 0.98 & 0.7 & 0.1  &  0.41 & 0.2 & 0.97 & 0.19 & 0.16 \\
 &  BLSTM  &  0.87 & 0.09 & 0.98 & 0.66 & 0.14  &  0.46 & 0.20 & 1.06 & 0.22 & 0.15 \\
\arrayrulecolor{lightgray} \hline \arrayrulecolor{black}
\multirow{2}*{ Muscle forces }
 & FFNN  &  0.85 & 0.29 & 0.99 & 0.07 & 0.08  &  0.37 & 0.17 & 0.69 & 0.10 & 0.18 \\
 &  LSTM  &  0.83 & 0.30 & 0.99 & 0.05 & 0.09  &  0.33 & 0.17 & 0.74 & 0.18 & 0.09 \\
\arrayrulecolor{lightgray} \hline \arrayrulecolor{black}
\multirow{2}*{ Muscle activations }
 & FFNN  &  0.88 & 0.25 & 1.0 & 0.21 & 0.05  &  0.29 & 0.15 & 0.62 & 0.08 & 0.07 \\
 &  LSTM  &  0.84 & 0.28 & 0.99 & 0.11 & 0.09  &  0.37 & 0.17 & 0.71 & 0.08 & 0.08 \\
\arrayrulecolor{lightgray} \hline \arrayrulecolor{black}

&  & \multicolumn{10}{c}{} \\[-0.8em]
& & \multicolumn{10}{c}{\textbf{Subject-naive}}  \\ 
&  & \multicolumn{10}{c}{} \\[-0.8em]
\arrayrulecolor{lightgray} \hline \arrayrulecolor{black}

\multirow{2}*{ Joint angles }
 & FFNN  &  0.86 & 0.17 & 0.98 & 0.32 & 0.07  &  0.82 & 0.65 & 2.42 & 0.20 & 0.60 \\
 &  BLSTM  &  0.86 & 0.13 & 0.98 & 0.56 & 0.09  &  1.10 & 1.02 & 3.56 & 0.20 & 0.73 \\
\arrayrulecolor{lightgray} \hline \arrayrulecolor{black}
\multirow{2}*{ Joint reaction forces }
 & FFNN  &  0.75 & 0.35 & 0.98 & -0.61 & 0.24  &  0.9 & 0.74 & 3.24 & 0.2 & 0.51 \\
 &  LSTM  &  0.64 & 0.34 & 0.92 & -0.24 & 0.31  &  0.84 & 0.77 & 3.19 & 0.15 & 0.51 \\
\arrayrulecolor{lightgray} \hline \arrayrulecolor{black}
\multirow{2}*{ Joint moments }
 & FFNN  &  0.85 & 0.12 & 0.99 & 0.56 & 0.18  &  1.12 & 1.02 & 3.24 & 0.25 & 0.72 \\
 &  GRU  &  0.75 & 0.13 & 0.94 & 0.5 & 0.20  &  0.89 & 0.68 & 2.50 & 0.35 & 0.22 \\
\arrayrulecolor{lightgray} \hline \arrayrulecolor{black}
\multirow{2}*{ Muscle forces }
 & FFNN  &  0.95 & 0.03 & 0.97 & 0.88 & 0.02  &  0.49 & 0.53 & 1.86 & 0.09 & 0.23 \\
 &  GRU  &  0.93 & 0.04 & 0.97 & 0.85 & 0.07  &  0.58 & 0.27 & 1.12 & 0.24 & 0.31 \\
\arrayrulecolor{lightgray} \hline \arrayrulecolor{black}
\multirow{2}*{ Muscle activations }
 & FFNN  &  0.93 & 0.04 & 0.96 & 0.84 & 0.04  &  0.42 & 0.28 & 0.96 & 0.10 & 0.30 \\
 &  LSTM  &  0.92 & 0.06 & 0.97 & 0.78 & 0.07  &  0.61 & 0.21 & 0.87 & 0.31 & 0.33 \\
\arrayrulecolor{lightgray} \hline \arrayrulecolor{black}
\hline
\end{tabular}
\caption{Average Pearson's correlation coefficient and Average NRMSE Values for Feed-Forward Neural Network (FFNN) and Recurrent Neural Network (RNN) predictions compared with Musculoskeletal (MSK) model outputs. Note: The average is taken over all output features and over all test trials (for a given output category). %IQR refers to inter-quartile range.
}
\label{tab:result2}
\end{center}
\end{table}

%%%%%%%%%%%%%%%%%%%%%%%%%%%%

\begin{table}[htb!]
\begin{center}
\begin{tabular}{ c | c | c c c c c }
 \hline
& & \multicolumn{5}{c}{RMSE} \\ 
 \hline
 Output & ML model  & Mean & SD  & Max & Min & IQR \\
\hline
\hline
&  & \multicolumn{5}{c}{} \\[-0.8em]
& &  \multicolumn{5}{c}{\textbf{Subject-exposed}}\\ 
&  & \multicolumn{5}{c}{} \\[-0.8em]
\arrayrulecolor{lightgray} \hline \arrayrulecolor{black}

\multirow{2}*{ Joint angles (degrees) }
 & FFNN  &  2.91 & 1.86 & 8.24 & 1.07 & 1.49 \\
 &  BLSTM  &  3.42 & 1.73 & 7.91 & 1.04 & 2.32 \\
\arrayrulecolor{lightgray} \hline \arrayrulecolor{black}
\multirow{2}*{ Joint reaction forces (\% Body Weight) }
 & FFNN  &  2.15 & 2.55 & 9.75 & 0.21 & 2.37 \\
 &  LSTM  &  2.08 & 2.37 & 7.84 & 0.24 & 2.38 \\
\arrayrulecolor{lightgray} \hline \arrayrulecolor{black}
\multirow{2}*{ Joint moments (\% Body Weight $\times$ Body Height ) }
 & FFNN  &  0.02 & 0.02 & 0.06 & 0.00 & 0.02 \\
 &  BLSTM  &  0.02 & 0.02 & 0.07 & 0.00 & 0.03 \\
\arrayrulecolor{lightgray} \hline \arrayrulecolor{black}
\multirow{2}*{ Muscle forces (\% Body Weight) }
 & FFNN  &  0.64 & 0.6 & 1.53 & 0.01 & 1.01 \\
 &  LSTM  &  0.45 & 0.37 & 0.97 & 0.03 & 0.64 \\
\arrayrulecolor{lightgray} \hline \arrayrulecolor{black}
\multirow{2}*{ Muscle activations (\%) }
 & FFNN  &  0.97 & 0.68 & 1.92 & 0.04 & 1.19 \\
 &  LSTM  &  1.38 & 1.05 & 2.78 & 0.04 & 1.85 \\
\arrayrulecolor{lightgray} \hline \arrayrulecolor{black}
&  & \multicolumn{5}{c}{} \\[-0.8em]
& & \multicolumn{5}{c}{\textbf{Subject-naive}}  \\ 
&  & \multicolumn{5}{c}{} \\[-0.8em]
\arrayrulecolor{lightgray} \hline \arrayrulecolor{black}
\multirow{2}*{ Joint angles (degrees) }
 & FFNN  &  9.07 & 3.54 & 16.49 & 4.31 & 5.21 \\
 &  BLSTM  &  11.09 & 4.62 & 20.37 & 3.40 & 6.07 \\
\arrayrulecolor{lightgray} \hline \arrayrulecolor{black}
\multirow{2}*{ Joint reaction forces (\% Body Weight) }
 & FFNN  &  5.52 & 6.74 & 26.57 & 0.28 & 6.00 \\
 &  LSTM  &  6.44 & 10.81 & 41.45 & 0.35 & 6.34 \\
\arrayrulecolor{lightgray} \hline \arrayrulecolor{black}
\multirow{2}*{ Joint moments (\% Body Weight $\times$ Body Height) }
 & FFNN  &  0.03 & 0.03 & 0.09 & 0.00 & 0.05 \\
 &  GRU  &  0.03 & 0.03 & 0.11 & 0.00 & 0.03 \\
\arrayrulecolor{lightgray} \hline \arrayrulecolor{black}
\multirow{2}*{ Muscle forces (\% Body Weight) }
 & FFNN  &  0.48 & 0.46 & 1.38 & 0.06 & 0.54 \\
 &  GRU  &  1.13 & 1.09 & 2.90 & 0.08 & 2.06 \\
\arrayrulecolor{lightgray} \hline \arrayrulecolor{black}
\multirow{2}*{ Muscle activations (\%) }
 & FFNN  &  1.09 & 0.65 & 2.03 & 0.17 & 1.02 \\
 &  LSTM  &  3.03 & 2.82 & 7.26 & 0.28 & 5.17 \\
\arrayrulecolor{lightgray} \hline \arrayrulecolor{black}
\hline
\end{tabular}
\caption{Average RMSE Values for Feed-Forward Neural Network (FFNN) and Recurrent Neural Network (RNN) predictions compared with Musculoskeletal (MSK) model outputs. Note: The average is taken over all output features and over all test trials (for a given output category). %
}
\label{tab:result3}
\end{center}
\end{table}

\end{document}